\documentclass[a4paper,fleqn]{cas-sc}
\pdfoutput = 1
\usepackage[authoryear,longnamesfirst]{natbib}
\usepackage{amsmath}
\usepackage{graphicx}
\usepackage{float}
\usepackage{subfigure}
\usepackage{color,xcolor}
\usepackage{lastpage}
\usepackage{multirow}
\usepackage{tabularx}
\usepackage{chngpage}
\usepackage{array}
\usepackage{booktabs}
\usepackage{apalike}
\usepackage{caption}
\usepackage{threeparttable}
\usepackage[title]{appendix}
\newcommand{\secref}[1]{Section~\ref{#1}}
\newcommand{\figref}[1]{Fig.~\ref{#1}}
\newcommand{\tabref}[1]{Table~\ref{#1}}
\newcommand{\appref}[1]{Appendix~\ref{#1}}
\def\tsc#1{\csdef{#1}{\textsc{\lowercase{#1}}\xspace}}
\tsc{WGM}
\tsc{QE}
\newtheorem{theorem}{Theorem}
\newtheorem{lemma}{Lemma}
\newdefinition{rmk}{Definition}
\newproof{pf}{Proof}
\newproof{pot}{Proof of Theorem \ref{thm}}

\begin{document}
\let\printorcid\relax
\def\floatpagepagefraction{1}
\def\textpagefraction{.001}

\shorttitle{FXAM: A Unified and Fast Interpretable Model for Predictive Analytics}    

\shortauthors{Yuanyuan Jiang et~al.}  

\title [mode = title]{FXAM: A Unified and Fast Interpretable Model for Predictive Analytics}  

%\tnotemark[<tnote number>] 

%\tnotetext[<tnote number>]{<tnote text>} 


\author[1,5]{Yuanyuan Jiang}

\ead{jyy_amy@163.com}

%\ead[url]{<URL>}

%\credit{<Credit authorship details>}

\author[2]{Rui Ding}
\cormark[1]

% Footnote of the second author
%\fnmark[b]
%\fntext[b]{Microsoft Research Center}
% Email id of the second author
\ead{juding@microsoft.com}

\cortext[cor1]{Corresponding author}

\author[3]{Tianchi Qiao}
\ead{tianchi-qiao@seu.edu.cn}

\author[4]{Yunan Zhu}
\ead{zhuyn@mail.ustc.edu.cn}

\author[2]{Shi Han}
\ead{shihan@microsoft.com}

\author[2]{Dongmei Zhang}
\ead{dongmeiz@microsoft.com}

\affiliation[1]{organization={School of Statistics, Renmin University of China},
            addressline={Haidian District}, 
            city={Beijing},
            postcode={100872}, 
            country={China}}
\affiliation[2]{organization={Microsoft Research Asia},
            addressline={Haidian District}, 
            city={Beijing},
            postcode={100080}, 
            country={China}}
\affiliation[3]{organization={School of Computer Science and Engineering, Southeast University},
            addressline={Nanjing}, 
            city={Jiangsu Province},
            postcode={211189}, 
            country={China}}
\affiliation[4]{organization={School of Information Science and Technology, University of Science and Technology of China},
            addressline={He Fei}, 
            city={Anhui Province},
            postcode={230031}, 
            country={China}}

\affiliation[5]{organization={Department of Machine Learning, Mohamed bin Zayed University of Artificial Intelligence},
            addressline={Masdar City}, 
            city={Abu Dhabi}, 
            country={United Arab Emirates}}

\begin{abstract}
Predictive analytics aims to build machine learning models to predict behavior patterns and use predictions to guide decision-making. Predictive analytics is human involved, thus the machine learning model is preferred to be interpretable. In literature, Generalized Additive Model (GAM) is a standard for interpretability. However, due to the one-to-many and many-to-one phenomena which appear commonly in real-world scenarios, existing GAMs have limitations to serve predictive analytics in terms of both accuracy and training efficiency. In this paper, we propose FXAM (Fast and eXplainable Additive Model), a unified and fast interpretable model for predictive analytics. FXAM extends GAM's modeling capability with a unified additive model for numerical, categorical, and temporal features. FXAM conducts a novel training procedure called Three-Stage Iteration (TSI). TSI corresponds to learning over numerical, categorical, and temporal features respectively. Each stage learns a local optimum by fixing the parameters of other stages. We design joint learning over categorical features and partial learning over temporal features to achieve high accuracy and training efficiency. We prove that TSI is guaranteed to converge to the global optimum. We further propose a set of optimization techniques to speed up FXAM's training algorithm to meet the needs of interactive analysis. Thorough evaluations conducted on diverse data sets verify that FXAM significantly outperforms existing GAMs in terms of training speed, and modeling categorical and temporal features. In terms of interpretability, we compare FXAM with the typical post-hoc approach XGBoost+SHAP on two real-world scenarios, which shows the superiority of FXAM's inherent interpretability for predictive analytics.
\end{abstract}


%\item 
%\item 
%\item 

% Keywords

\maketitle

\section{Introduction}\label{sec1}

\begin{figure}[htbp]
\centering{\includegraphics[width=0.6\textwidth]{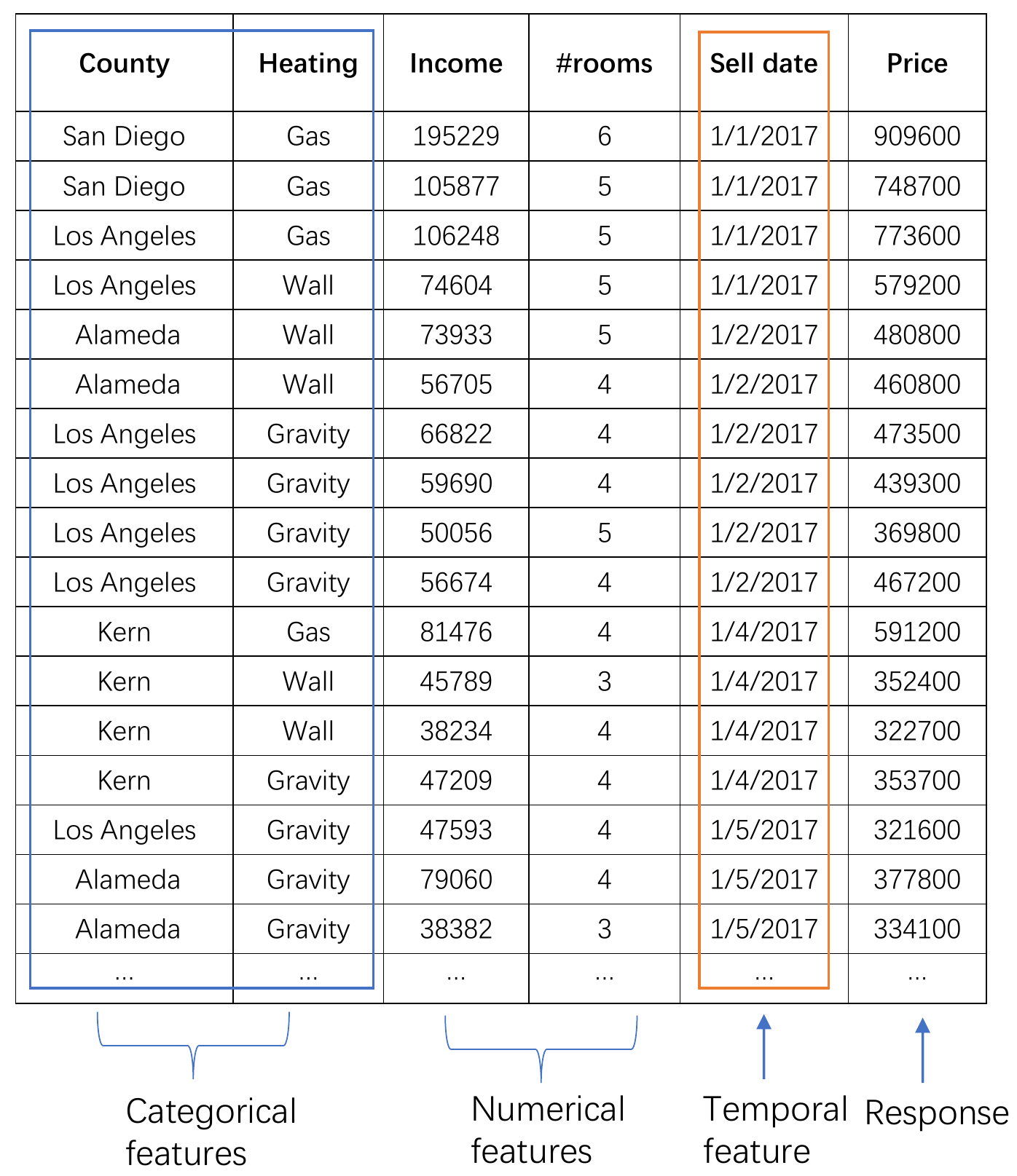}}
\caption{An example of the multi-dimensional data set.}\label{Figure 1}
\end{figure}

Expert systems are often used in decision-making scenarios~\citep{zimm:1987}, especially in the high-stakes domains~\citep{meske:2022,simkute:2021} (such as healthcare, criminal justice, or finance) where they can provide valuable insights and recommendations to help with complex decision-making processes. Predictive analytics is an essential topic in expert systems~\citep{Cheng:2018} and aims to predict behavior patterns from multi-dimensional data and use predictions to guide decision-making~\citep{Fin:2014, KRam:2021}. Multi-dimensional data is conceptually organized in a tabular format that consists of a set of records, where each record is represented by a set of attributes, with one attribute called response (i.e., the target to be predicted) and the others called features (or predictors), which are used to predict the response. A multi-dimensional data set typically consists of three types of features: numerical, categorical, and temporal. \figref{Figure 1} shows an example of a house sale data set with several features, such as $Income$ (numerical), $County$ (categorical), $Sell date$ (temporal), etc., and the response is $Price$. By building an ML model from multi-dimensional data, follow-up analysis is performed, such as understanding existing records or predicting response on a newly unseen record. 

Predictive analytics is human-involved and is frequently conducted for high-stakes prediction applications thus the ML model is preferred to be interpretable~\citep{Rud:2019}. In the literature, the Generalized Additive Model (GAM) is a standard for interpretability~\citep{HasTib:1990}. GAM untangles the overall prediction by summing up contributions from each feature (before applying the link function), thus retaining interpretability. Moreover, GAM's training procedure (a.k.a. backfitting) works by iterative smoothing of partial residuals over each feature, which guarantees convergence to an optimal solution (when suitable smoothers are chosen). GAMs are continuously being developed, such as GA2M ~\citep{Lou:2013}, GAMut~\citep{Hoh:2019}, multi-class GAM~\citep{Zhang:2019}, ReluctantGAM~\citep{TayTib:2020}, COGAM~\citep{Abd:2020}, etc. However, due to the {\em one-to-many} and {\em many-to-one} phenomena that appear commonly in multi-dimensional data, existing GAMs have limitations in serving predictive analytics. 

{\bf One-to-many}: Learning multiple components from each temporal feature. A numerical feature typically introduces a locally smoothing constraint on its contribution to response, but a temporal feature (e.g., `Sell date') introduces multiple global constraints from a time-series perspective: it is desirable to identify multiple components from a temporal feature, such as monthly repeating (i.e., seasonality) component, long-term progression pattern (i.e., trend), or aperiodic cycles~\citep{ZOzy:2006}, etc. However, existing GAMs treat a temporal feature as an ordinary numerical feature and thus only learn a single smoothing component. As a result, their model capacity is limited w.r.t. dealing with temporal features.

{\bf Many-to-one}: Since there is no local smoothing constraint across categorical values, users focus on identifying the contribution of each distinct value (e.g., the extra cost of buying a house when it is located in `County = LA'). Existing GAMs conduct histogram-type smoothing per categorical feature, which converges slowly since only the weights of values of a specific categorical feature are updated in each iteration, while all the other weights (w.r.t. distinct values from other categorical features) are fixed. If the weights of values across all categorical features could be updated simultaneously, we could speed up model training.

Moreover, predictive analytics is often conducted iteratively. Fast training makes the analysis more interactive and continuous, which cannot be easily facilitated by existing GAMs due to their unsatisfactory training speed. To address these challenges, we propose FXAM: a unified, fast, and interpretable model for predictive analytics. FXAM has significant advantages in the following areas: 

{\bf Modeling}. FXAM extends GAM's modeling capability with a unified additive model for numerical, categorical, and temporal features. For each temporal feature, FXAM identifies multiple components in terms of trend and seasonality; FXAM proposes a homogeneous set to model categorical values across all categorical features and represents each value via one-hot encoding.

{\bf Training}. FXAM conducts a novel training procedure called Three-Stage Iteration (TSI). The three stages correspond to learning over numerical, categorical, and temporal features, respectively. Each stage learns a local optimum by fixing the parameters of other stages. Specifically, we design joint learning over categorical features and partial learning over temporal features to achieve high training efficiency and high accuracy. We also provide theoretical analysis in Theorem~\ref{thm1} to show that TSI converges to a global optimum.

{\bf Efficiency}. We further propose two optimization techniques (i.e., intelligent sampling and dynamic feature iteration) with theoretical guidance to speed up FXAM's training algorithm to meet the needs of interactive analysis. 

In summary, we make the following contributions:
    \begin{itemize}
      \item FXAM extends GAMs modeling capability with a unified model for numerical, categorical, and temporal features.
      \item We propose FXAM's training procedure: Three Stage Iteration, and prove its convergence and optimality.
      \item We propose two optimization techniques to speed up FXAM's training algorithm.
      \item We conduct evaluations and verify that FXAM significantly outperforms existing GAMs in terms of training speed and modeling categorical and temporal features.
    \end{itemize}
% Use the style of numbering in square brackets.
% If nothing is used, default style will be taken.
%\begin{enumerate}[a)]
%\item 
%\item 
%\item 
%\end{enumerate}  

%\begin{itemize}
%\item 
%\item 
%\item 
%\end{itemize}  

%\begin{description}
%\item[]
%\item[] 
%\item[] 
%\end{description}  

% Figure

\section{Related Work}\label{sec2}

{\bf Predictive analytics \& iML (interactive Machine Learning)}. Predictive analytics is often conducted for high-stakes prediction applications, such as healthcare, finance, or phishing detection thus the ML model is preferred to be interpretable~\citep{Rud:2019}. Operationally, predictive analytics is often conducted iteratively and interactively, thus iML (interactive Machine Learning) is becoming a cornerstone for predictive analytics~\citep{Fols:2003, Abd:2018}, which requires ML model to respond in an interactive fashion. Therefore, ML model's training efficiency becomes primarily important.

{\bf XAI (Explainable artificial intelligence)}. XAI is becoming a hot topic~\citep{Lom:2006, Mil:2019, Kau:2020} and current XAI techniques can generally be grouped into two categories~\citep{DLHu:2019, Arrieta:2020}. {\bf Interpretable}: designing inherently explainable ML models~\citep{Lou:2013, Caruana:2015, Jung:2017} or {\bf Explainable}: providing post-hoc explanations to opaque models~\citep{Ribeiro:2016, Lundberg:2017, Tan:2018}, depending on the time when explainability is obtained~\citep{Molnar:2020}. In the domain of predictive analytics, interpretable ML models tend to be more useful since explainability is needed throughout the analysis process, such as probing different subsets of data, incorporating domain constraints, or understanding model mechanisms locally or globally. FXAM is an extension of GAM, thus retaining interpretability.

{\bf GAMs}. GAMs are gaining great attention in the literature of interpretable machine learning~\citep{Rud:2019,Arrieta:2020,Chang:2021,Linardatos:2021}, mainly due to its standard for interpretability~\citep{Wang:2021} and its broad adoptions in the real world~\citep{Pierrot:2011,Calabrese:2012,Wang:2021,Tomic:2014}. GAM-based approaches are continuously being developed: GA2M~\citep{Lou:2013} models pairwise feature interaction; multi-class GAM~\citep{Zhang:2019} generalizes GAM to the multi-class setting; COGAM~\citep{Abd:2020} and ReluctantGAM~\citep{TayTib:2020} impose linear constraints on certain features to achieve a tradeoff between cognitive load and model accuracy. There also exists work on modeling GAM's shape functions by neural nets such as NAM~\citep{Agarwal:2020}and GAMI-Net~\citep{Yang:2021}.

FXAM is complementary to these works by modeling numerical, categorical, and temporal features in a unified way and by proposing an efficient and accurate training procedure. In FXAM, joint learning is conducted over all categorical features instead of per-feature learning (e.g., histogram-type smoothing in pyGAM) to improve training efficiency; partial learning is adopted to accurately learn trend and seasonality components from each temporal feature. Such an approach can be naturally extended to learn arbitrary components. Although there exists work on identifying seasonality components by adopting cyclic cubic spline, they require additional efforts on data preprocessing~\citep{Simpson:2014}, and the learned seasonal component is restricted to be identical in each period thus progressive changes of seasonal component (e.g., amplifying or damping) cannot be captured. Lastly, such a preprocessing approach is difficult to extend to learn other components, such as aperiodic cyclic components~\citep{Hyndman:2011,Hyndman:2018}. 	

\section{Terms and Notations}\label{sec3}
Except for special instructions, we use uppercase italics for variables, uppercase bold letters for matrices, lowercase bold letters for vectors, lowercase letters for scalars, subscripts for the variable index, and superscripts with parentheses for the instance index. Our discussion will center on a response random variable Y, and $p$ numerical features $X_{1},  \ldots, X_{p}$; $q$ categorical features $Z_{1}, \ldots, Z_{q}$; $u$ temporal features $T_{1},  \ldots, T_{u}$. Given a multi-dimensional data set $\mathcal{D}$ consists of $N$ instances, the realizations of these random variables can be denoted by $(y^{(1)}, x_{1}^{(1)}, \ldots, x_{p}^{(1)}, z_{1}^{(1)},  \ldots, z_{q}^{(1)}, t_{1}^{(1)}, \ldots, t_{u}^{(1)} ), \ldots, ( y^{(N)}$, $x_{1}^{(N)}, \ldots,  x_{p}^{(N)}, z_{1}^{(N)}, \ldots, z_{q}^{(N)}, t_{1}^{(N)},  \ldots, t_{u}^{(N)} )$. The summary of terms and notations is shown in \tabref{tabl1}.

\begin{table*}[width=0.9\textwidth]
\centering
\caption{A summary of terms and notations.}
\label{tabl1}
\begin{tabular}{m{0.2\textwidth} m{0.36\textwidth} m{0.3\textwidth}} 
\hline
\multicolumn{1}{l}{\textbf{Type}} & \multicolumn{1}{l}{\textbf{Symbol}} & \multicolumn{1}{l}{\textbf{Explanation}} \\ 
\hline
\multirow{2}{*}{Data set} & $\mathcal{D}$ & Data set \\ 
\cline{2-3}
 & $N$ & Data set size \\ 
\hline
\multirow{2}{*}{Response} & $Y$ & Random variable \\ 
\cline{2-3}
& $\boldsymbol{y}^{T} = \left(y^{(1)}, y^{(2)} \ldots, y^{(N)} \right)$ & Instances\\  
\hline
\multirow{3}{*}{Predictors} & $X_1, X_2, \ldots, X_p$ & $p$ numerical features \\ 
\cline{2-3}
 & $Z_1, Z_2, \ldots, Z_q$ & $q$ categorical features \\ 
\cline{2-3}
 & $T_1, T_2, \ldots, T_u$ & $u$ temporal features \\ 
\hline
\multirow{12}{*}{Categorical features} & $dom(Z_m)$, $m=1,\ldots,q$ & The set including the distinct values for the categorical feature $Z_m$  \\ 
\cline{2-3}
 & $H_{cat}=\bigcup_{m=1}^{q} dom\left(Z_m\right)$ & The homogenous set including the distinct values for all categorical features \\ 
\cline{2-3}
 & $c=|H_{cat}|$ & Total cardinality over all categorical features \\ 
\cline{2-3}
 & $(O_1,O_2,\ldots,O_c)$ where $O_j \in \left\{0,1\right\}$, \quad \quad \quad $j=1, \ldots, c$ & The $q$-hot vector representing the encoding of $Z_1, \ldots, Z_q$ \\
 \cline{2-3}
 & $f_Z\left(O_j\right) = {\beta_j O_j}$ & The parameterized form by representing categorical values $Z_1, \ldots, Z_q$ in the $c$-dimensional vector\\ 
\hline
\multirow{3}{*}{Numerical features} & $H_{num}^i$, $i=1,\ldots,p$ & The Hilbert space of the measurable function $f_i(X_i)$ over numerical feature $X_i$  \\ 
\cline{2-3}
& $f_i(X_i)$  & The univariate smooth function modeling the contributions of $X_i$, $f_i(X_i) \in H_{num}^i$ \\ 
\hline
\multirow{18}{*}{Temporal features} & $H_{tem}^k$, $k=1,\ldots,u$ & The Hilbert space of the measurable functions $f_{S_k}(T_k)$ for seasonality and $f_{T_k}(T_k)$ for trend over temporal feature $T_k$ \\ 
\cline{2-3}
 & $f_{S_k}(T_k)$ & The function modeling the seasonality of $T_k$, $f_{S_k}(T_k) \in H_{tem}^k$ \\ 
\cline{2-3}
 & $f_{T_k}(T_k)$ & The function modeling the trend of $T_k$,  $f_{T_k}(T_k) \in H_{tem}^k$ \\ 
\cline{2-3}
 & $d_k$ & The period of seasonal component, $d_k > 1$ and $d_k$ is a hyperparameter; \\ 
\cline{2-3}
 & $dom(T_k) = \{t_k^{(1)}, \ldots, t_k^{(N)}\}$ & The set including all the ordered values of $T_k$ in $\mathcal{D}$, where $t_k^{(1)} \leq \ldots \leq t_k^{(N)}$\\ 
\cline{2-3}
 & $t_k^{(l)}-t_k^{(l-1)} = 0$ or $\tau$ & The corresponding gap between two consecutive time points, $\tau$ is a constant, $l = 2, \ldots, N$\\ 
\cline{2-3}
 & $\mathcal{T}_{k, \varphi}:=\left\{t_k^{(l)} \mid t_{k}^{(l)} / {\tau} \bmod d_k=\varphi\right\}$ & The set of time points with phase-$\varphi$, where $\varphi \in\{1, \cdots, d_k-1\}$, $l = 1, \ldots, N$ \\ 
\cline{2-3}
 & $dom(T_k)=\bigcup_{\varphi=0}^{d_k - 1} \mathcal{T}_{k, \varphi}$ & $dom(T_k)$ can be written as the sum of $\mathcal{T}_{k,\varphi}$ \\
\hline
\end{tabular}
\end{table*}

{\bf Categorical features}. For each $m\in{1,\ldots,q}$, denote the domain of $Z_{m}$ as $dom(Z_{m})$, which indicates the distinct values for categorical feature $Z_{m}$. For instance, each element in $dom(Z_{m})$ can be a string value that is composed of the specific value in $Z_{m}$ with the corresponding feature name as the suffix. Hence, the domains of different categorical features are disjoint. 

\begin{figure}[htbp]
\centering{\includegraphics[width=0.6\textwidth]{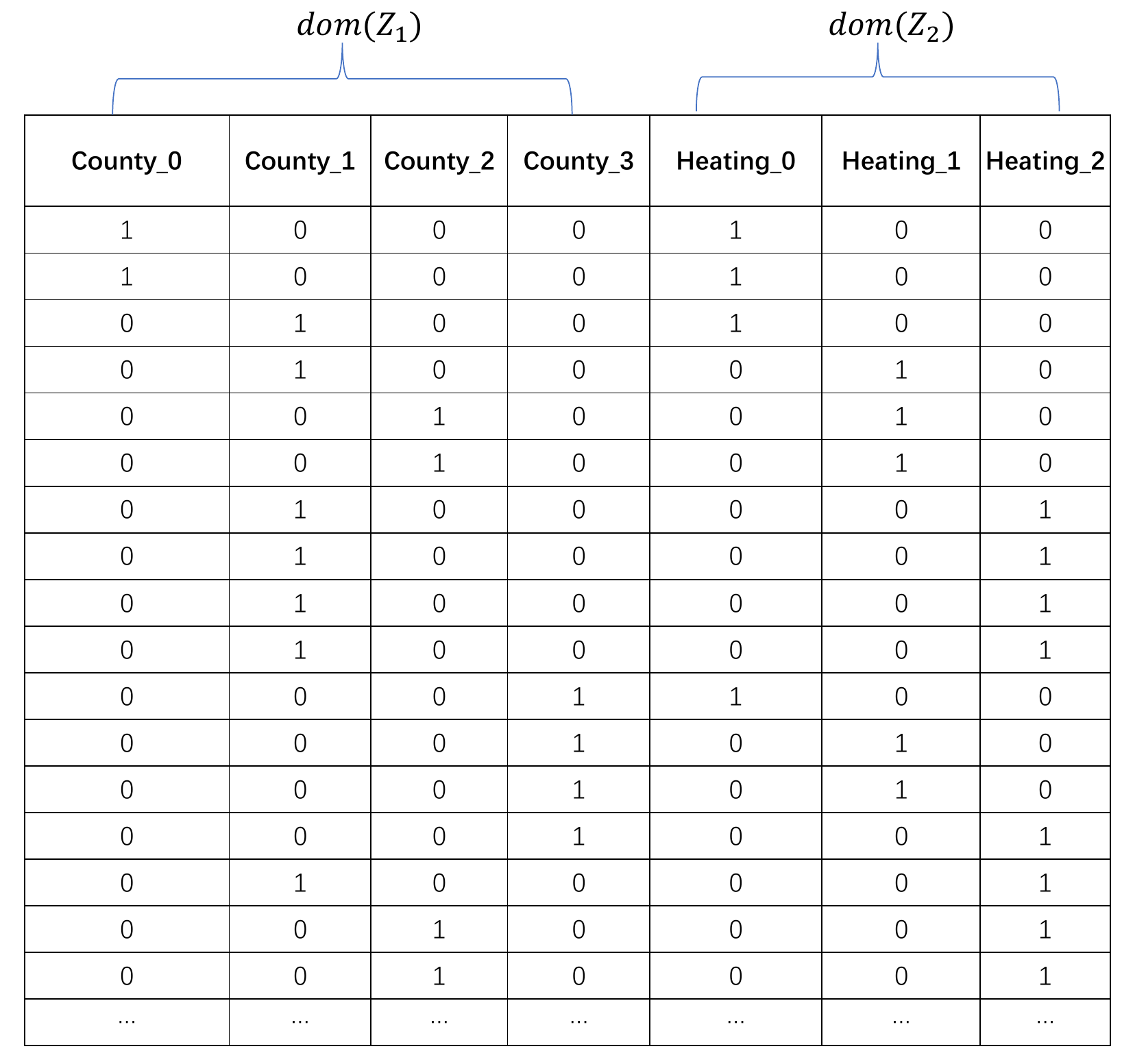}}
\caption{An example of the processed categorical variables.}\label{Figure 2}
\end{figure}

Denote $H_{cat}=\bigcup_{m=1}^{q} dom\left(Z_{m}\right)$ as the homogenous set, and $c=|H_{cat}|$ as total cardinality (i.e., number of distinct values) over all categorical features. Denote $O_{j} \in\{0,1\}$, $j=1,2, \ldots,c$, thus any instantiation of $Z_{1}, \ldots, Z_{q}$ can be represented by a unique  $q$-hot vector $\left(O_{1}, \ldots, O_{c}\right)$ provided that pre-specified indices are assigned to elements in $H_{cat}$. Continuing with the example in \figref{Figure 1}, the categorical variables can be processed as shown in \figref{Figure 2}, where each row represents the $q$-hot vector, and the $j$th column represents the variable $O_j$.

{\bf Numerical features}. Following standard convention, for each $i\in{1,\ldots,p}$, let $\mathcal{H}_{num}^{i}$ denote the Hilbert space of measurable functions $f_i(X_i)$ such that $E\left[f_i\right]=0,\ E\left[f_i^2\right]<\infty$ and inner product $\left \langle f_i,f_i' \right \rangle=E\left[f_if_i'\right]$. Here the expectation is defined over the probability density distribution corresponding to the training data. For our purpose, we would like to learn (or estimate) a shape function $f_i(X_i) \in H_{num}^{i} $ for each numerical feature.

{\bf Temporal features}.  To identify trend component, for each $k\in{1,\ldots,u}$, let $\mathcal{H}_{tem}^{k}$ denote the Hilbert space of measurable function $f_{T_k}(T_k)$ for trend and $f_{S_k}(T_k)$ for seasonality over temporal feature $T_k$. $\mathcal{H}_{tem}^{k}$ is with the same property as $\mathcal{H}_{num}^{i}$. To identify the seasonal component, denote the period of the seasonal component as a positive integer $d_k>1$. Note that $d_k$ is an input parameter based on domain knowledge, which is common practice in the business data analytics domain~\citep{Cleveland:1990, Wen:2019}.

\begin{figure*}[htbp]
\centering{\includegraphics[width=0.8 \textwidth]{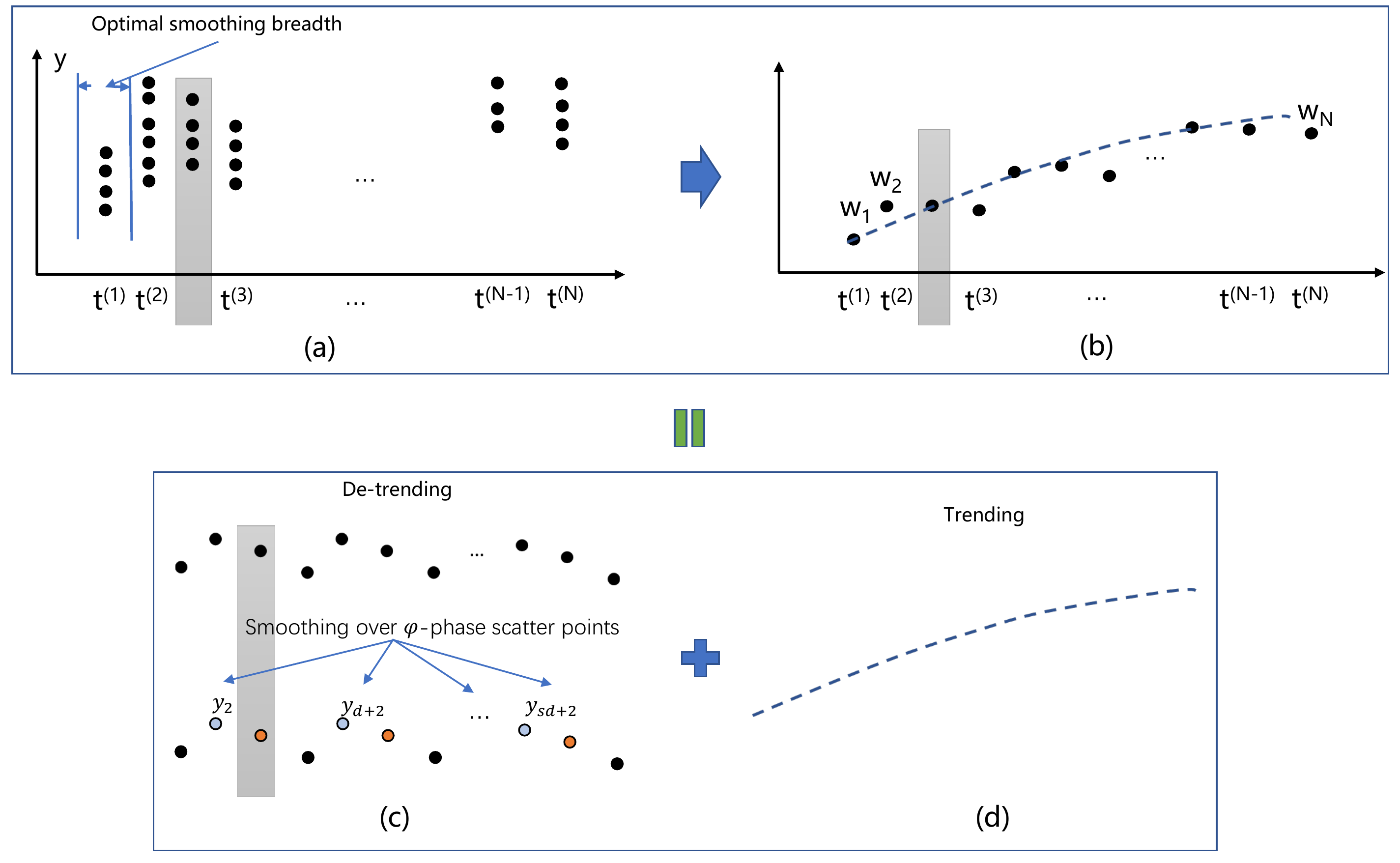}}
\caption{An example of the processed temporal variables.}\label{Figure 3}
\end{figure*}

We order the values of $T_k$ in $\mathcal{D}$ as: $t_{k}^{(1)}\le\ldots\le t_{k}^{(N)}$, and assume $\left \{t_{k}^{(l)}-t_{k}^{(l-1)}|l=2,\ldots,N \right\} = \left \{0,\tau \right\} $ for the sake of simplicity: Continuing with the example in \figref{Figure 1}, it is common that there would be multiple instances with the same value on $t_{k}^{(*)}$ as shown in \figref{Figure 3}(a). Instances with the same value at time $t_{k}^{(*)}$ are compressed into one instance with weight $w_*$ as shown in \figref{Figure 3}(b). Thus the corresponding gap between two consecutive time points $t_{k}^{(l)}-t_{k}^{(l-1)} = 0$ or $t_{k}^{(l)}-t_{k}^{(l-1)} = \tau$ allows us to treat $t_{k}^{(\ast)}$ as discrete time points. Now it is easy to decompose the time series of temporal feature $T_k$ into a trend $f_{T_k}$ and a seasonality $f_{S_k}$ as shown in \figref{Figure 3}(c) and \figref{Figure 3}(d). Denote $\mathcal{T}_{k, \varphi}:=\left\{t_{k}^{(l)} \mid t_{k}^{(l)} / \tau \bmod d_k=\varphi, \forall l\right\}$ as phase-$\varphi$ set since all the elements in $\mathcal{T}_{k,\varphi}$ share same phase $\varphi\in\left\{0,1,\ldots,d_k-1\right\}$. As shown in \figref{Figure 3}(c), instances of the same color belong to the same set $\mathcal{T}_{k,\varphi}$. It is easy to see that $\mathcal{T}_{k,i}$, $\mathcal{T}_{k,j}$ ($i,j \in\left\{0,1,\ldots,d_k-1\right\}$ and $i \neq j$) are mutually disjoint and thus $\left\{t_{k}^{(1)},\ldots,t_{k}^{(N)} \right\}=\mathcal{T}_{k,0}+\ldots+\mathcal{T}_{k,d_k-1}$. Our approach can also easily deal with missing data as the shaded part shown in \figref{Figure 3} (i.e., the gap between two consecutive time points could be larger than $\tau$ in a data set due to insufficient sample), which is discussed in \secref{sec4.5}.

\section{Approach}\label{sec4}
Without loss of generality, we focus on the case where link function $g(Y)$ is an identity function, thus we are focusing on the regression problem subsequently. We first illustrate the FXAM's modeling over numerical, categorical, and temporal features, and then we propose FXAM's training procedure called Three-Stage Iteration. At last, two optimization techniques are presented to further improve FXAM's training efficiency.

\subsection{FXAM's Modeling}
We model FXAM as follows:

\begin{gather}\label{eq1}
\begin{aligned}
	E\left(Y|X_1, \ldots, X_p; Z_1, \ldots, Z_q; T_1, \ldots, T_u\right) &=\sum_{i=1}^{p}{f_i\left(X_i\right)}+\sum_{j=1}^{c}{f_Z\left(O_j\right)} \\
& + \sum_{k=1}^{u}\left[f_{T_k}\left(T_k\right)+f_{S_k}\left(T_k\right)+other\_component\left(T_k\right)\right]
\end{aligned}
\end{gather}

Here $f_Z(O_j)=\beta_jO_j$, $j=1, \ldots, c$. $\beta_j\in\mathbb{R}$ is the parameter and $f_i\in\mathcal{H}_{num}^{i}$, $f_{T_k}\in\mathcal{H}_{tem}^{k}$, $f_{S_k}\in\mathcal{H}_{tem}^{k}$ are the functions we want to learn. $O_j\in\left\{0,1 \right\}$ is obtained by one-hot encoding over homogenous set $H_{cat}$. The overall model is composed of three parts additively w.r.t. modeling numerical, categorical, and temporal features respectively.

{\bf Modeling numerical features}. Following standard convention, $f_i$ is the shape function that models the contribution of $X_i$ (w.r.t. response $Y$) by a univariate smooth function.

{\bf Modeling categorical features}. We conduct one-hot encoding for each element in the homogeneous set $H_{cat}$. Specifically, $\sum_{j=1}^{c}f_Z\left(O_j\right) = \sum_{j=1}^{c}{\beta_j O_j}$ is a parameterized form by representing categorical values $Z_1, \ldots, Z_q$ in a $c$-dimensional $q$-hot vector and assigns a weight $\beta_j$ to each entry $O_j$. 

{\bf Modeling temporal features}. $\left[f_{T_k}\left(T_k\right)+f_{S_k}\left(T_k\right)+\ldots\right]$ explicitly decomposes the time series of temporal feature $T_k$ into a trend $f_{T_k}$, a seasonality $f_{S_k}$ and some other signals. Such decomposition expresses multiple components from a single feature to address the one-to-many phenomenon. 

For the sake of simplicity, we focus on seasonal and trend decomposition, and we assume $u=1$ (i.e., only one temporal feature) in subsequent illustrations. We thus drop the subscript $k$ and use $T$ to denote the temporal feature. Note that the theorem of FXAM's convergence is valid for arbitrary $u$ and FXAM's training procedure can be easily extended to support multiple temporal features (details are elaborated in \secref{sec4.5}).

\subsection{FXAM's Optimization} 
The objective function we want to minimize is:

\begin{gather}\label{eq2}
\begin{aligned}	\mathcal{L}\left(f_{1}, \ldots, f_{p}, \beta_1, \ldots, \beta_c, f_{T}, f_{S}\right)  &= \sum_{l=1}^{N}\left(y^{(l)}-\sum_{i=1}^{p} f_{i}\left(x_{i}^{(l)}\right)-\sum_{j=1}^{c} f_Z\left(o_{j}^{(l)}\right)-f_T\left(t^{(l)}\right)-f_S\left(t^{(l)}\right)\right)^{2}  \\
& + \lambda \sum_{i=1}^{p} J\left(f_{i}\right)+\lambda_{Z} 
\boldsymbol{\beta}^{T} \boldsymbol{\beta}+\lambda_{T} J(f_T)+\lambda_{S} \sum_{\varphi=0}^{d-1} J\left(f_{S_{\varphi}}\right)
\end{aligned}
\end{gather}

Here $\boldsymbol{\beta} ^ {T} = (\beta_1,\ldots,\beta_c)$. Eq.~\eqref{eq2} consists of the total square error and the other regularization items. Functional $J(f):=\int\left[f^{\prime \prime}(v)\right]^{2} d v$ thus $\lambda J\left(f\right)$ trades off the smoothness of $f$ with its goodness-to-fit. In addition to standard regularization for numerical features $\lambda J\left(f_{i}\right)$ and trend $\lambda_{T} J(f_T)$, we divide seasonal component $f_S$ into $d$ sub-components $f_{S_{\varphi}}$ ($f_S:=f_{S_{0}} \oplus \ldots \oplus f_{S_{d-1}}$ indicates that overall seasonal component $f_{S}$ which domain-merges all the sub-components $f_{S_{\varphi}}$) and apply regularization per $f_{S_{\varphi}}$. By doing so, we impose smoothness for each phase-equivalent sub-component $f_{S_\varphi}$, which is helpful to convey the overall repeating pattern. We propose standard $L_2$ regularization $\lambda_Z \boldsymbol{\beta}^T\boldsymbol{\beta}$ correspondingly.

{\bf Quadratic Form of Objective Function}. By standard calculus~\citep{Reinsch:1967}, the optimal solution for minimizing a square error with regularization $\lambda J\left(f_i\right)$ is natural cubic spline smoothing with knots at $x_{i}^{(1)},\ldots,x_{i}^{(N)}$, thus the vector version of objective function $\mathcal{L}$ can be expressed as a quadratic form: 

\begin{gather}\label{eq3}
\begin{aligned}
\mathcal{L}\left(\boldsymbol{f}_{1}, \ldots, \boldsymbol{f}_{p}, \boldsymbol{f}_{Z}, \boldsymbol{f}_{T}, \boldsymbol{f}_{S}\right)  &= \left\|\boldsymbol{y}-\sum_{i=1}^{p} \boldsymbol{f}_{i}-\boldsymbol{f}_{Z}-\boldsymbol{f}_{T}-\boldsymbol{f}_{S}\right\|^{2}  \\
& + \lambda \sum_{i=1}^{p} \boldsymbol{f}_{i}^{T} \boldsymbol{K}_{i} \boldsymbol{f}_{i}+\lambda_{Z} \boldsymbol{\beta}^{T} \boldsymbol{\beta}+\lambda_{T} \boldsymbol{f}_{T}^{T} \boldsymbol{K}_{T} \boldsymbol{f}_{T} + \lambda_{S} \sum_{\varphi=0}^{d-1} \boldsymbol{f}_{S_{\varphi}}^{T} \boldsymbol{K}_{S_{\varphi}} \boldsymbol{f}_{S_{\varphi}}
\end{aligned}
\end{gather}

In Eq.~\eqref{eq3}, $\|*\|^{2}$ denotes the total square error and we use $\boldsymbol{f}_i, \boldsymbol{f}_Z ,\ \boldsymbol{f}_T$ and $\boldsymbol{f}_S\in\mathbb{R}^N$ as the vector version realizations of $f_i, f_Z, f_T, f_S$ in Eq.~\eqref{eq2} respectively. Here $\boldsymbol{f}_Z = \boldsymbol{Z} \boldsymbol{\beta}$, $\boldsymbol{Z}$ is a $N \times c$ design matrix corresponding to $N$ $q$-hot encoded vectors from categorical features as shown in~\figref{Figure 2}. $\boldsymbol{y}^{T}=(y^{(1)}, \ldots, y^{(N)})$, $\boldsymbol{f}_i^{T}= \left(f_i({x_i^{(1)}}), \ldots, f_i({x_i^{(N)}}) \right)$, $\boldsymbol{f}_T^{T}=\left(f_T(t^{(1)}), \right.$ $\left.\ldots, f_T(t^{(N)})\right)$, $\boldsymbol{f}_S^{T}=\left(f_S(t^{(1)}), \ldots, f_S(t^{(N)})\right)$.  $\boldsymbol{K}_i$ is a $N\times N$ matrix pre-calculated by values $x_{i}^{(1)},\ldots,x_{i}^{(N)}$~\citep{Buja:1989}. $\boldsymbol{K}_T$ is calculated the same way. $\boldsymbol{K}_{S_\varphi}$ is an $N\times N$ matrix obtained by applying cubic spline smoother over $\mathcal{T}_\varphi$ and then re-ordering the indices of records with a permutation matrix $\boldsymbol{P}_\varphi$. Specifically, $\boldsymbol{K}_{S_\varphi}=\boldsymbol{P}_\varphi^T\left[\begin{matrix} \widetilde{\boldsymbol{K}_{S_\varphi}}&0\\0&0\\\end{matrix}\right]\boldsymbol{P}_\varphi$, where $\widetilde{\boldsymbol{K}_{S_\varphi}}$ is a $\left|\mathcal{T}_\varphi\right|\times\left|\mathcal{T}_\varphi\right|$ matrix w.r.t. cubic spline smoothing over knots $\left \{t_{\varphi}^{(1)},t_{\varphi}^{(2)},\ldots,t_\varphi^{(\left|\mathcal{T}_\varphi\right|)} \right \}$ (i.e. $\mathcal{T}_\varphi$), and $\boldsymbol{P}_\varphi^T$ is a $N\times N$ permutation matrix mapping the indices of these knots into the original indices of elements in $T$.

{\bf Analysis of Optimality}. To minimize $\mathcal{L}$ in Eq.~\eqref{eq3}, we derive $\mathcal{L}$'s stationary solution via FXAM's normal equations:

\begin{gather*}
\begin{aligned}
&\nabla_{\boldsymbol{f}_{i}} \mathcal{L}=\left.0\right|_{i: 1, \ldots, p} \\
&\nabla_{\boldsymbol{\beta}} \mathcal{L}=0 \\
&\nabla_{\boldsymbol{f}_{T}} \mathcal{L}=0 \\
&\nabla_{\boldsymbol{f}_{S_{\varphi}}} \mathcal{L}=\left.0\right|_{\varphi: 0, \ldots, d-1}
\end{aligned} \quad \Rightarrow\left[\begin{array}{cccccc}
\boldsymbol{I} & \boldsymbol{M}_{Z} & \boldsymbol{M}_{Z} & \boldsymbol{M}_{Z} & \ldots & \boldsymbol{M}_{Z} \\
\boldsymbol{M}_{1} & \boldsymbol{I} & \boldsymbol{M}_{1} & \boldsymbol{M}_{1} & \ldots & \boldsymbol{M}_{1} \\
\boldsymbol{M}_{2} & \boldsymbol{M}_{2} & \boldsymbol{I} & \boldsymbol{M}_{2} & \ldots & \boldsymbol{M}_{2} \\
\vdots & \vdots & \vdots & \ddots & \vdots & \vdots \\
\boldsymbol{M}_{T} & \boldsymbol{M}_{T} & \boldsymbol{M}_{T} & \ldots & \boldsymbol{I} & \boldsymbol{M}_{T} \\
\boldsymbol{M}_{S} & \boldsymbol{M}_{S} & \boldsymbol{M}_{S} & \ldots & \boldsymbol{M}_{S} & \boldsymbol{I}
\end{array}\right]\left[\begin{array}{c}
\boldsymbol{f}_{Z} \\
\boldsymbol{f}_{1} \\
\boldsymbol{f}_{2} \\
\vdots \\
\boldsymbol{f}_{T} \\
\boldsymbol{f}_{S}
\end{array}\right]=\left[\begin{array}{c}
\boldsymbol{M}_{Z} \boldsymbol{y} \\
\boldsymbol{M} _{1} \boldsymbol{y} \\
\boldsymbol{M}_{2} \boldsymbol{y} \\
\vdots \\
\boldsymbol{M}_{T} \boldsymbol{y} \\
\boldsymbol{M}_{S} \boldsymbol{y}
\end{array}\right]
\end{gather*}

where
\begin{gather} \label{eq4}
\left\{\begin{array}{l}
\boldsymbol{M}_{Z}=\boldsymbol{Z}\left(\boldsymbol{Z}^{T} \boldsymbol{Z}+\lambda_{Z} \boldsymbol{I}\right)^{-1} \boldsymbol{Z}^{T} \\
\boldsymbol{M}_{i}=\left(\boldsymbol{I}+\lambda \boldsymbol{K}_{i}\right)^{-1}, \forall i \in\{1, \ldots, p\} \\
\boldsymbol{M}_{T}=\left(\boldsymbol{I}+\lambda_{T} \boldsymbol{K}_{T}\right)^{-1} \\
\boldsymbol{M}_{S}=\boldsymbol{P}^{T}\left[\begin{array}{ccc}
\left(\boldsymbol{I}+\lambda_{S} \widetilde{\boldsymbol{K}_{S_{0}}}\right)^{-1} & \ldots & 0 \\
\vdots & \ddots & \vdots \\
0 & \ldots & \left(\boldsymbol{I}+\lambda_{S} \widetilde{\boldsymbol{K}_{S_{d-1}}}\right)^{-1}
\end{array}\right] \boldsymbol{P}
\end{array}\right.
\end{gather}

In Eq.~\eqref{eq4}, $\boldsymbol{P}=\boldsymbol{P}_{0} \ldots \boldsymbol{P}_{d-1}$ is the overall permutation matrix, by mapping indices of elements from $\left\{t^{(1)}, \ldots, t^{(N)}\right\}$ to the indices of elements in $\left\{\mathcal{T}_{0}, \ldots, \mathcal{T}_{d-1}\right\}$. Then estimate $\boldsymbol{f}_{1}, \ldots, \boldsymbol{f}_{p}, \boldsymbol{f}_{Z}, \boldsymbol{f}_{T}, \boldsymbol{f}_{S}$ by given $\boldsymbol{M}_{1}, \ldots, \boldsymbol{M}_{p}, \boldsymbol{M}_{Z}, \boldsymbol{M}_{T}, $ \\ $ \boldsymbol{M}_{S}~\text{and}~ \boldsymbol{y}  $. By definition, a solution of FXAM's normal equations is a local optimum of $\mathcal{L}$, below we further prove that the solution also achieves the global optimum of $\mathcal{L}$.

\begin{theorem}\label{thm1}
Solutions of FXAM's normal equations exist and are the global optimum
\end{theorem}

\begin{pf}
According to theorem 2 in \citet{Buja:1989}, the solutions of normal equations exist and are globally optimal if each smoothing matrix $\boldsymbol{M}_{i}, \boldsymbol{M}_{Z}, \boldsymbol{M}_{T}$, or $\boldsymbol{M}_{S}$ is symmetric and shrinking (i.e., with eigenvalues in $[0,1]$ ). Thus we check $\boldsymbol{M}_{i}, \boldsymbol{M}_{Z}, \boldsymbol{M}_{T}$, and $\boldsymbol{M}_{S}$ one by one:

$\boldsymbol{M}_{i}, \boldsymbol{M}_{T}$ are indeed symmetric and shrinking according to standard analysis of cubic spline smoothing matrix.

Re-write $\boldsymbol{M}_{Z}=\boldsymbol{Z} \boldsymbol{A} \boldsymbol{Z}^{T}$ where $\boldsymbol{A}=\left(\boldsymbol{Z}^{T} \boldsymbol{Z}+\lambda_{Z} \boldsymbol{I}\right)^{-1}$. It is easy to see that $\boldsymbol{A}$ is a symmetric matrix thus $\boldsymbol{M}_{Z}^{T}=\left(\boldsymbol{Z}^{T}\right)^{T} \boldsymbol{A}^{T} \boldsymbol{Z}^{T}=\boldsymbol{Z} \boldsymbol{A} \boldsymbol{Z}^{T}=\boldsymbol{M}_{Z}$.

Denote singular value decomposition of $\boldsymbol{Z}$ is $\boldsymbol{Z}=\boldsymbol{U} \boldsymbol{\Lambda} \boldsymbol{V}^{T}$, where $\boldsymbol{U}$ and $\boldsymbol{V}$ are orthogonal matrices, $\boldsymbol{\Lambda}$ is a $c \times c$ diagonal matrix, with diagonal entries $\Lambda_{11} \geq \cdots \geq \Lambda_{c c} \geq 0$. Thus we have $\boldsymbol{M}_{Z} \boldsymbol{y}=\boldsymbol{Z}\left(\boldsymbol{Z}^{T} \boldsymbol{Z}\!+\!\lambda \boldsymbol{I}\right)^{-1} \boldsymbol{Z}^{T} \boldsymbol{y}=\boldsymbol{U} \boldsymbol{\Lambda} \boldsymbol{V}^{T}\!\left(\boldsymbol{V} \boldsymbol{\Lambda}^{2} \boldsymbol{V}^{T}\!+\!\lambda \boldsymbol{I}\right)^{-1} \boldsymbol{V} \boldsymbol{\Lambda} \boldsymbol{U}^{T} \boldsymbol{y}=\boldsymbol{U} \boldsymbol{\Lambda} \boldsymbol{V}^{T}\!\left(\boldsymbol{V} \boldsymbol{\Lambda}^{2} \boldsymbol{V}^{T}\!+\!\lambda \boldsymbol{V} \boldsymbol{I} \boldsymbol{V}^{T}\right)^{-1} \boldsymbol{V} \boldsymbol{\Lambda} \boldsymbol{U}^{T} \boldsymbol{y}= \boldsymbol{U} \boldsymbol{\Lambda} \boldsymbol{V}^{T} \boldsymbol{V}\left(\boldsymbol{\Lambda}^{2}
\!+\!\lambda \boldsymbol{I} \right)^{-1} \boldsymbol{V}^{-1} \boldsymbol{V} \boldsymbol{\Lambda} \boldsymbol{U}^{T} \boldsymbol{y}$
$=\boldsymbol{U} \boldsymbol{\Lambda} \left(\boldsymbol{\Lambda}^{2} \!+\!\lambda \boldsymbol{I}\right)^{-1} \boldsymbol{\Lambda} \boldsymbol{U}^{T} \boldsymbol{y}= \sum_{j=1}^{c} \boldsymbol{u_{j}} \frac{\Lambda_{jj}^{2}}{\Lambda_{jj}^{2}\!+\!\lambda} \boldsymbol{u_{j}}^{T} \boldsymbol{y}$, thus the eigenvalues $\frac{\Lambda_{jj}^{2}}{\Lambda_{jj}^{2}\!+\!\lambda}$ are in $[0,1]$ considering $\lambda>0$.

Re-write $\boldsymbol{M}_{S}=\boldsymbol{P}^{T} \boldsymbol{\Theta} \boldsymbol{P}$. Since each $\widetilde{\boldsymbol{K}_{S_\varphi}}$ is a symmetric matrix, thus $\left(\boldsymbol{I}+\lambda_{S} \widetilde{\boldsymbol{K}_{S_\varphi}}\right)^{-1}$ is symmetric and $\boldsymbol{\Theta}$ is symmetric, thus $\boldsymbol{M}$ is symmetric. Due to the shrinking property of $\left(\boldsymbol{I}+\lambda_S \widetilde{\boldsymbol{K}_{S_\varphi}}\right)^{-1}$, and considering $\boldsymbol{\Theta}$ is a block-diagonal matrix with $\left(\boldsymbol{I}+\lambda_{S} \widetilde{\boldsymbol{K}_{S_\varphi}}\right)^{-1}$ as its blocks, thus $\boldsymbol{\Theta}$ is also shrinking: $\|\boldsymbol{\Theta} \boldsymbol{y}\|^{2} \leq\|\boldsymbol{y}\|^{2}, \forall \boldsymbol{y}$. So $\left\|\boldsymbol{M}_{S} \boldsymbol{y}\right\|^{2}=\boldsymbol{y}^{T} \boldsymbol{M}_{S}^{T} \boldsymbol{M}_{S} \boldsymbol{y}=$ $\boldsymbol{y}^{T} \boldsymbol{P}^{T} \boldsymbol{\Theta}^{T} \boldsymbol{\Theta} \boldsymbol{P}\boldsymbol{y}=\| \boldsymbol{\Theta} \boldsymbol{P} \boldsymbol{y}\|^{2} \leq\|\boldsymbol{P} \boldsymbol{y}\|^{2}=\|\boldsymbol{y}\|^{2}$, thus $\boldsymbol{M}_{S}$ is shrinking. \hfill  $\blacksquare$
\end{pf}

\subsection{FXAM's Training}
\begin{figure}[htbp]
\centering{\includegraphics[width=0.8\textwidth]{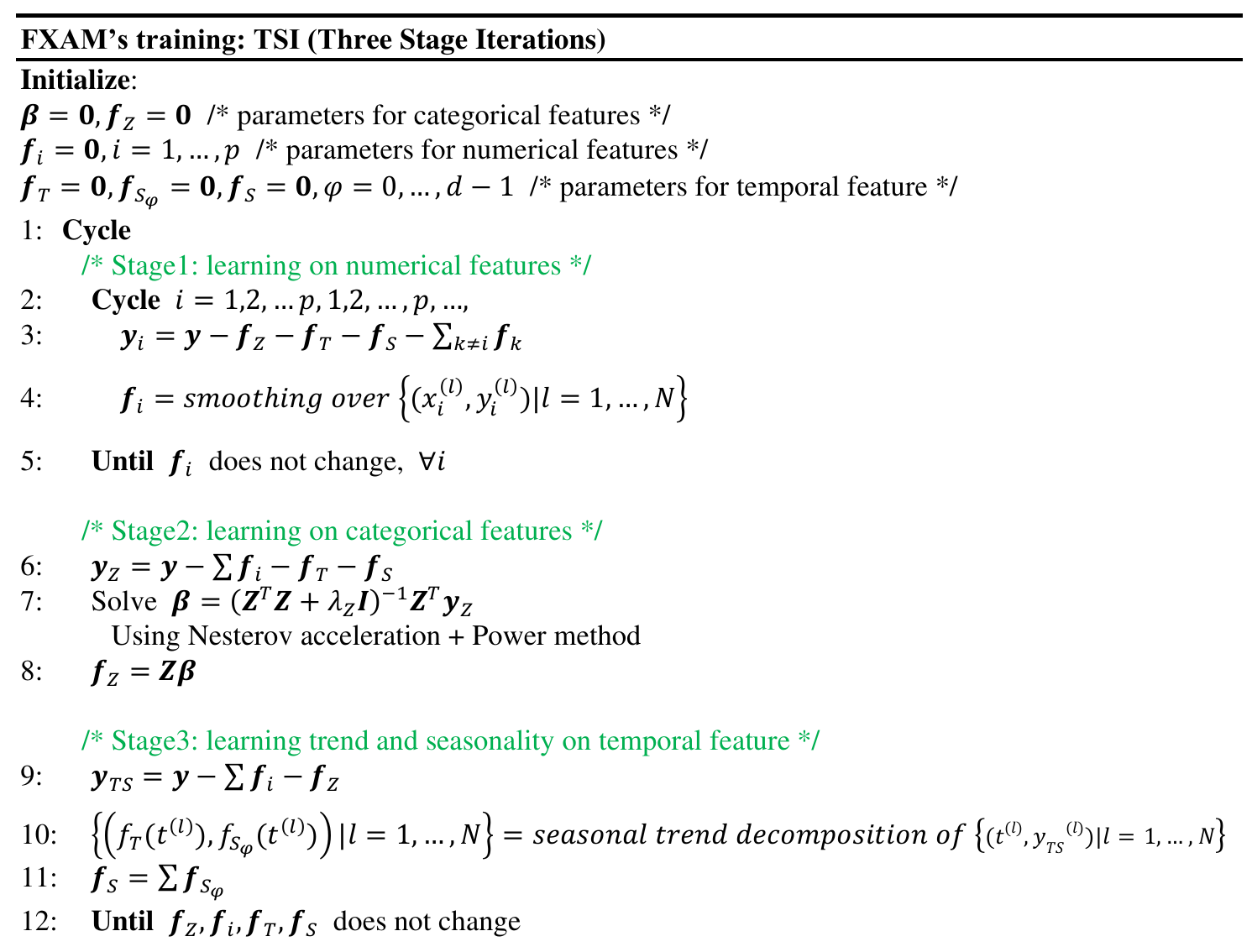}}
\caption{Three Stage Iteration procedure.}\label{Figure 4}
\end{figure}

To solve FXAM's normal equations, we extend backfitting and develop a novel training procedure: Three-Stage Iteration (TSI). TSI consists of three stages: learning over numerical, categorical, and temporal features, respectively. As shown in \figref{Figure 4}: standard backfitting is applied over numerical features (line $2 - 5$), we additionally design {\em joint learning} over categorical features (line $6 - 8$ ) to improve training efficiency, and {\em partial learning} over temporal features (line $9 - 11$) to learn trend and seasonal components to improve model accuracy. TSI is more appealing in that it maintains convergence to the solution of FXAM's normal equations, thus the output of TSI is the global optimum of $\mathcal{L}$.

\begin{figure}[ht]
\centering{\includegraphics[width=0.8\textwidth]{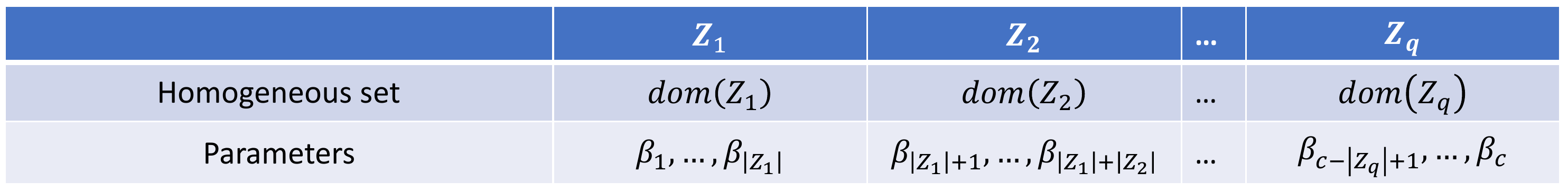}}
\caption{Joint learning vs. per-feature learning (Joint learning: all parameters $\beta_{1}, \ldots, \beta_{c}$ are jointly updated in each iteration. Per-feature learning: only parameters of $Z_{i}$ are updated in each iteration).}\label{Figure 5}
\end{figure}

{\bf Joint learning on categorical features}. To deal with categorical features, existing GAMs conduct per-feature smoothing (e.g., histogram-type smoothing in pyGAM), which converges slowly since only the weights of a specific categorical feature (e.g., $Z_{1}$) are updated (e.g., $\beta_{1}, \ldots, \beta_{\left|Z_{1}\right|}$) but all the other weights (e.g., $\beta_{\left|Z_{1}\right|+1}, \ldots, \beta_{c}$) are fixed in each iteration (depicted in \figref{Figure 5}). In contrast, we pull all categorical values into a homogeneous set $H_{cat}$ and learn all the parameters $\beta_{1}, \ldots, \beta_{c}$ jointly. Joint learning enables accelerating gradient descent via adopting improved momentum: we adopt Nesterov acceleration and power method~\citep{Nesterov:1983, Sutskever:2013} to improve training efficiency (line 7 in \figref{Figure 4}). In our experiment, joint learning achieves $3 - 10$ times faster than per-feature learning.

Nesterov acceleration can be viewed as an improved momentum, thus making convergence significantly faster than gradient descent especially when the cardinality of the homogenous set is large~\citep{Nesterov:1983,Sutskever:2013}. As depicted in line 7 of \figref{Figure 4}, the task of learning the contributions of categorical values boils down to calculating $\boldsymbol{\beta}=\left(\boldsymbol{Z}^{T} \boldsymbol{Z}+\right.$ $\left.\lambda_{Z} \boldsymbol{I} \right)^{-1}  \boldsymbol{Z}^{T} \boldsymbol{y}_{Z}$. We adopt Nesterov's Gradient Acceleration (NGA) to estimate $\boldsymbol{\beta}$ together with power iteration to identify optimal learning rate $\mu$. The adoption of NGA with power iteration in our problem is depicted on the left side of \figref{Figure 7}. The optimal learning rate $\mu$ equals the greatest eigenvalue of $\boldsymbol{Z}^{T} \boldsymbol{Z}+\lambda_{Z} \boldsymbol{I}$, which can be efficiently identified by power iteration. The complexity of our algorithm is $O\left(k c^{2}\right)$ where $k$ is \#iterations of NGA that $k \ll c$. Note that directly calculating matrix inversion has complexity $O\left(c^{3}\right)$ which is unaffordable when $c$ is large.

\begin{figure}[htbp]
\centering{\includegraphics[width=0.8\textwidth]{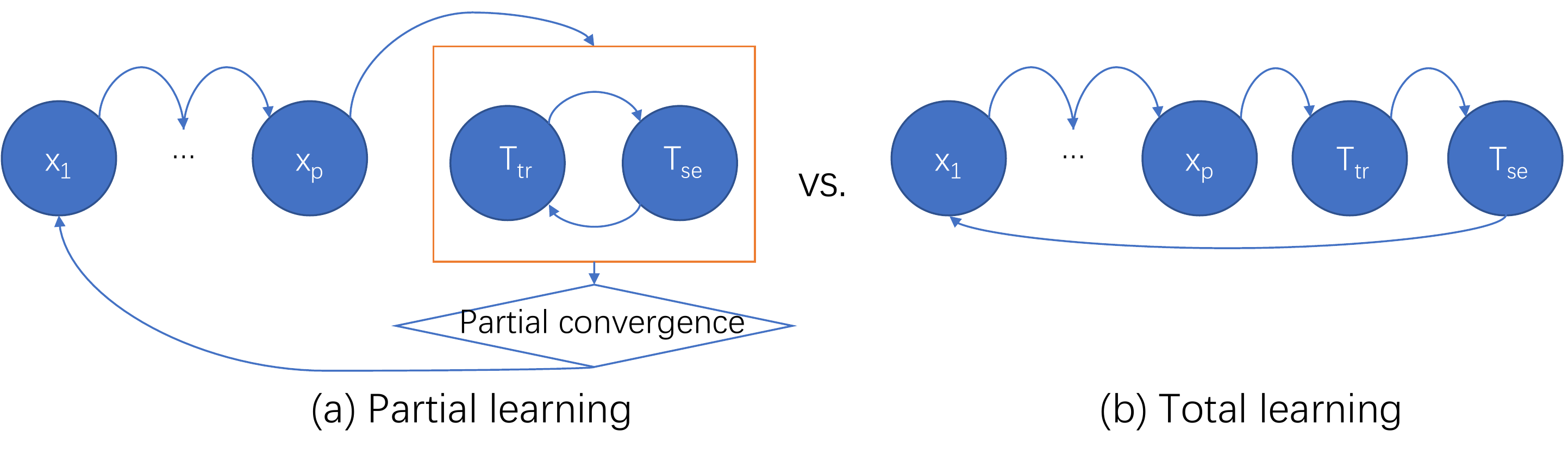}}
\caption{Partial learning vs. total learning.}\label{Figure 6}
\end{figure}

\begin{figure*}[htbp]
\centering{\includegraphics[width=0.8\textwidth]{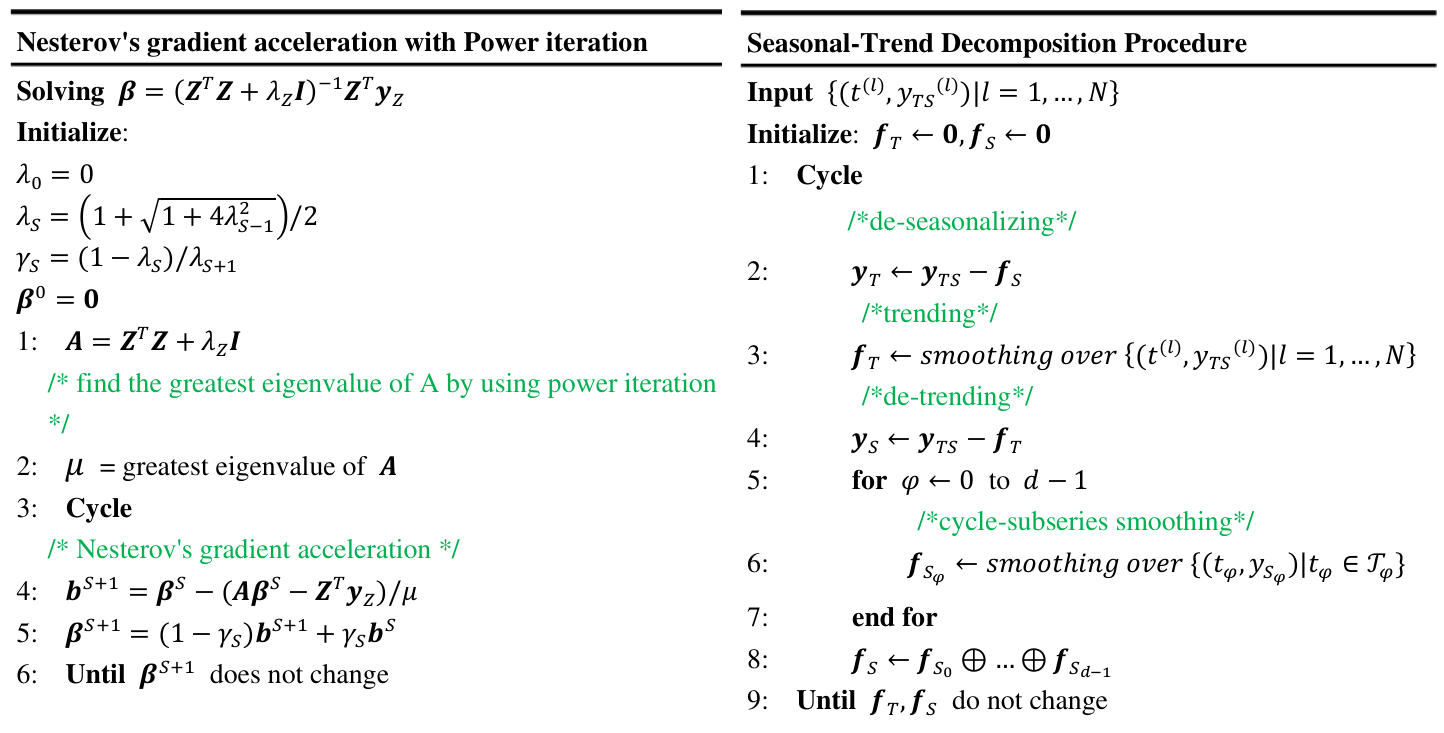}}
\caption{Details of Training in TSI (Left part: Nesterov’s acceleration. Right part: Seasonal-Trend Decomposition).}\label{Figure 7}
\end{figure*}

{\bf Partial learning on temporal features}. We adopt partial learning to accurately learn multiple components from each temporal feature $T$. Specifically, we first duplicate $T$ into two virtual features $T_\text {tr}$ and $T_\text{se}$, and then apply smoother $\boldsymbol{M}_{T}$ (de-trend operation) on $T_\text{tr}$ and $\boldsymbol{M}_{S}$ (de-seasonal operation) on $T_{\text {se}}$ iteratively until a partial convergence, and then move out to other features (\figref{Figure 6}(a)). In contrast, total learning (i.e., standard approach) puts $T_\text{tr}$ and $T_\text{se}$ with numerical features together and conducts backfitting (\figref{Figure 6}(b)) without partial convergence, which could lead to undesirable entanglement: the learned trend component exhibits small periodicity (i.e., carries partial seasonal component), and the learned seasonal component exhibits slow drift (i.e., carries partial trend component). In comparison, partial learning learns more accurate trends and seasonal components.

The details about learning multiple components (i.e., seasonal trend decomposition) have been shown in \figref{Figure 7}: we conduct local iteration to identify trend $\boldsymbol{f}_T$ and seasonality $\boldsymbol{f}_S$ from temporal feature $T$ as depicted on the right side of \figref{Figure 7}. $\boldsymbol{f}_T$ is obtained by applying smoothing matrix $\boldsymbol{M}_{T}$ (line 3) and cycle-subseries smoothing (line 6) is applied to smoothing matrix $\boldsymbol{M}_{S_{\varphi}}$ to obtain $\boldsymbol{f}_{S_{\varphi}}$. Such local iteration ensures effective decomposition of $\boldsymbol{f}_T$ and $\boldsymbol{f}_S$, leading to more stable and accurate results. According to Theorem \ref{thm2}, such local iteration still preserves overall convergence.

\begin{theorem} \label{thm2}
    TSI converges to a solution of FXAM's normal equations.
\end{theorem}

\begin{pf}
A full cycle of TSI can be written as a linear map $\boldsymbol{\mathcal{K}}=\left(\boldsymbol{\Phi}_{S} \boldsymbol{\Phi}_{T}\right)^{\infty} \boldsymbol{\Phi}_{Z} \boldsymbol{K}^{\infty}$, where $\boldsymbol{K}^{\infty}=\left(\boldsymbol{\Phi}_{p} \boldsymbol{\Phi}_{p-1} \ldots \boldsymbol{\Phi}_{1}\right)^{\infty}$. Here $\boldsymbol{K}^{\infty}$ represents partial convergence of numerical features (Stage1), and $\left(\boldsymbol{\Phi}_{S} \boldsymbol{\Phi}_{T}\right)^{\infty}$ represents partial convergence of one temporal feature $t$ (Stage3). We need to prove $\boldsymbol{\mathcal{K}}^{m}$ converges to $\boldsymbol{\mathcal{K}}^{\infty}$, and $\boldsymbol{\mathcal{K}}^{\infty}$ is a solution of FXAM's normal equations.

Denote the index set $\mathcal{I}:=\{1,2, \ldots, p, Z, T, S\}$. We only need to prove that TSI converges to a solution of FXAM's homogenous equations (i.e., FXAM's normal equations with $\boldsymbol{y}=\boldsymbol{0}$) because a general solution is a solution of homogenous equations plus an arbitrary solution of FXAM's normal equations. Denote the loss function of homogenous equations as $\mathcal{L}_{0}(\boldsymbol{f}):=\left\|\sum_{j \in \mathcal{I}} \boldsymbol{f}_{j}\right\|^{2}+\sum_{j \in \mathcal{I}} \boldsymbol{f}_{j}^{T}\left(\boldsymbol{M}_{j}^{-}-\boldsymbol{I}\right) \boldsymbol{f}_{j}$.

We define a linear map $\boldsymbol{\Phi_{j}}$ to describe the updating of $j$th component in TSI when $\boldsymbol{y}=\boldsymbol{0}$:
$$
\boldsymbol{\Phi_{j}}:\left[\begin{array}{c}
\boldsymbol{f}_{1} \\
\vdots \\
\boldsymbol{f}_{p} \\
\boldsymbol{f}_{Z} \\
\boldsymbol{f}_{T} \\
\boldsymbol{f}_{S}
\end{array}\right] \equiv \boldsymbol{f} \rightarrow\left[\begin{array}{c}
\boldsymbol{f}_{1} \\
\vdots \\
-\boldsymbol{M}_{j} \sum_{i \in \mathcal{I}, i \neq j} \boldsymbol{f}_{i} \\
\boldsymbol{f}_{Z} \\
\boldsymbol{f}_{T} \\
\boldsymbol{f}_{S}
\end{array}\right], \forall j \in \mathcal{I}
$$

A full cycle of backfitting over numerical features is then described by $\boldsymbol{K}=$ $\boldsymbol{\Phi}_{p} \boldsymbol{\Phi}_{p-1} \ldots \boldsymbol{\Phi}_{1}$. Denote the $m$ full cycles as $\boldsymbol{K}^{m}$. It is obvious that $\boldsymbol{K}^{m}$ converges to a limit $\boldsymbol{K}^{\infty}$ (we can view this as a standard task of backfitting over pure numerical features) therefore with property $\boldsymbol{K} \boldsymbol{K}^{\infty}=\boldsymbol{K}^{\infty}$. Note that $\boldsymbol{K}^{\infty}$ describes the procedure of stage 1, thus the full cycle of the entire TSI is $\boldsymbol{\mathcal{K}}=\boldsymbol{\Phi}_{S} \boldsymbol{\Phi}_{T} \boldsymbol{\Phi}_{Z} \boldsymbol{K}^{\infty}$. Since each component of $\boldsymbol{\mathcal{K}}$ is minimizer of $\mathcal{L}_{0}(\boldsymbol{f})$ and since $\mathcal{L}_{0}$ is a quadratic form, hence $\mathcal{L}_{0}(\boldsymbol{\mathcal{K}} \boldsymbol{f}) \leq \mathcal{L}_{0}(\boldsymbol{f})$. When $\mathcal{L}_{0}(\boldsymbol{\mathcal{K}} \boldsymbol{f})=\mathcal{L}_{0}(\boldsymbol{f})$, no strict descent is possible on any component, thus $\boldsymbol{\Phi}_{S} \boldsymbol{f}=\boldsymbol{f}, \boldsymbol{\Phi}_{T} \boldsymbol{f}=$ $\boldsymbol{f}, \boldsymbol{\Phi}_{Z} \boldsymbol{f}=\boldsymbol{f}, \boldsymbol{K^{\infty}} \boldsymbol{f}=\boldsymbol{f}$. Considering $\boldsymbol{K} \boldsymbol{K}^{\infty}=\boldsymbol{K}^{\infty}$, thus $\boldsymbol{K} \boldsymbol{K}^{\infty} \boldsymbol{f}=\boldsymbol{K}^{\infty} \boldsymbol{f} \Leftrightarrow \boldsymbol{K} \boldsymbol{f}=\boldsymbol{f}$ when descent vanishes. Since each component $\boldsymbol{\Phi}_{j}$ of $\boldsymbol{K}$ only updates separate $\boldsymbol{f}_{j}$, thus $\boldsymbol{K} \boldsymbol{f}=\boldsymbol{f} \Leftrightarrow \boldsymbol{\Phi}_{j} \boldsymbol{f}=\boldsymbol{f}, \forall j \in\{1, \ldots, p\}$. So descent vanishes on any $\boldsymbol{f}$ equivalent to $\boldsymbol{\Phi}_{j} \boldsymbol{f}=\boldsymbol{f}, \forall j \in \mathcal{I}$. Meanwhile, such $\boldsymbol{f}$ satisfies homogenous equations, which indicates $\mathcal{L}_{0}(\boldsymbol{f})=$ 0 according to theorem 5 in \citet{Buja:1989}. In summary, we have a linear mapping $\boldsymbol{\mathcal{K}}$ satisfying $\mathcal{L}_{0}(\boldsymbol{\mathcal{K}} \boldsymbol{f})<\mathcal{L}_{0}(\boldsymbol{f})$ when $\mathcal{L}_{0}(\boldsymbol{f})>0$ and $\boldsymbol{\mathcal{K}} \boldsymbol{f}=\boldsymbol{f}$ when $\mathcal{L}_{0}(\boldsymbol{f})=0$. According to theorem 8 of \citet{Buja:1989}, $\boldsymbol{\mathcal{K}}^{m}$ converges to $\boldsymbol{\mathcal{K}}^{\infty}$.   \hfill  $\blacksquare$
\end{pf}

\subsection{Improving Training Efficiency}
To further improve training speed, we propose two techniques: intelligent sampling and dynamic feature iteration to improve backfitting efficiency of Stage 1 in TSI. Last but not least, we adopt a recent fast version of kernel smoothing instead of standard cubic spline smoothing. 

{\bf Intelligent Sampling}. For large-scale data sets (e.g., $\mathrm{N}>$ $100,000)$, we estimate the shape function over a sample set as better initialization of $\boldsymbol{f}_{i}$ (Initialize part in \figref{Figure 4}). By doing so, the number of iterations towards convergence could be reduced while the time cost of sampling-based initialization could be negligible if the sample size is chosen appropriately. Next, we illustrate how to determine an appropriate sample size $n$.

Considering the task of smoothing over $\{(x^{(j)},y^{(j)})| j=1, \ldots, N \}$. Assume the records are drawn from ground-truth function $F(X): y^{(i)}=F(x^{(i)})+\epsilon^{(i)}$ where $\epsilon^{(i)}$ are i.i.d. random errors with $E\left(\epsilon^{(i)}\right)=0, \operatorname{Var}\left(\epsilon^{(i)}\right) \leq \sigma^{2}$. Denote $U$ as maximum slope of $F$, i.e., $\left|F\left(x^{(i)}\right)-F\left(x^{(j)}\right)\right| \leq U\left\|x^{(i)}-x^{(j)}\right\|, \forall x^{(i)}, x^{(j)}$ (Lipschitz condition). Denote $f_{N}$ and $f_{n}$ as shape functions obtained by smoothing over $\{(x^{(j)},y^{(j)})| j=1, \ldots, \textit{data set size} \}$ on full set and sample set respectively, we verify that the sample variation $E\left\|f_{N}-f_{n}\right\|^{2}$ has an upper bound according to Lemma~\ref{lema1}. Therefore the sampling error is controllable and the estimates are accurate.

\begin{lemma}\label{lema1}
$E\left\|f_{N}-f_{n}\right\|^{2} \leq 4 c\left[\left(\sigma^{2}+\sup F^{2}\right) U / n\right]^{2 / 3}$
\end{lemma}

\begin{pf}
    Sample variation is the difference between a smoothing function $f_{N}(X)$ obtained from all records and another smoothing function $f_{n}(X)$ obtained from sampled records with sample size $n$.

According to theorem $5.2$ in \citet{Gyorfi:2002}, for any kernel smoother $f_{N}, E\left\|f_{N}-F\right\|^{2} \leq c\left(\frac{\left(\sigma^{2}+\sup F^{2}\right) U}{N}\right)^{2 / 3}$, $\forall N$. This provides a way to estimate the upper bound of sample variation by 

\begin{gather*}
\begin{aligned}
E\left\|f_{N}-f_{n}\right\|^{2} &= \int\left(f_{N}(X)-f_{n}(X)\right)^{2} \mu(d X) \\
&=\int\left(f_{N}(X)-F(X)+F(X)-f_{n}(X)\right)^{2} \mu(d X) \\
&\leq \int\left(f_{N}(X)-F(x)+F(X)-f_{n}(X)\right)^{2} \mu(d X) \\
&+\int\left(f_{N}(X)-F(X)-F(X)+f_{n}(X)\right)^{2} \mu(d X) \\
&=2\left(E\left\|f_{N}-F\right\|^{2}+E\left\|f_{n}-F\right\|^{2}\right) \\
&\leq 4 E\left\|f_{n}-F\right\|^{2} \\
&=4 c\left(\frac{\left(\sigma^{2}+\sup |F|^{2}\right) U}{n}\right)^{\frac{2}{3}} \hspace{25em} \blacksquare
\end{aligned}
\end{gather*}  
\end{pf}

According to Lemma \ref{lema1}, sample variation depends on sample size $n$, noise level $\sigma$, maximum slope $U$ and square of maximum absolute value $\sup F^{2}$ of $F$. To bound sample variation for all features, sample size $n_{i}$ for feature $X_{i}$ should be $n_{i} \propto\left(\sigma_{i}^{2}+\sup F_{i}^{2}\right) U_{i}$. We use a pre-specified small sample size $n_{0}$ (e.g., 10,000) to approximately estimate $F_{i} \approx$ $f_{n_{0}}$ for feature $X_{i}$, and then use it to further obtain estimation of $\sigma_{i}$ and $U_{i}$.

We define $n^{*}=\underset{i}{\max} \gamma\left(\sigma_{i}^{2}+\sup F_{i}^{2}\right) U_{i}$ as sample size applied to initialization for all numerical features, which is conservative since under sample size $n^{*}$, sample variations for all features are bounded. $\gamma$ is a hyperparameter.

{\bf Dynamic Feature Iteration (DFI)}. DFI is a heuristic algorithm. We propose DFI to dynamically adjust the order of features for smoothing. Smoothing over a feature with higher predictive power will reduce more loss locally (i.e., within the current cycle) thus achieving faster convergence. We propose a lightweight estimator to calculate and update the predictive power of each feature and use it to dynamically order features.

\begin{rmk}
The predictive power of $X_i$ is defined as $Power_{i}=2TSS \cdot r_{i}^{2} /(N-2)-\left(2 \hat{U}_{i} Bh\right)^{2}$
\end{rmk}
Here $TSS=\sum_{l=1}^{N}\left(\widetilde{y}^{(l)}-\bar{\tilde{y}}\right)^{2}$, $\widetilde{y}^{(l)}$ is current partial residual and $\bar{\tilde{y}}$ is its average. Assume $\tilde{y}^{(l)}$ is the $l$-th instance of variable $\tilde{Y}$, thus $r_{i}$ is the Pearson correlation coefficient of $X_{i}$ and $\tilde{Y}$. $B$ is the bounded support of kernel $K_{h}$, and $h$ is corresponding smoothing bandwidth. $\hat{U}_{i}$ is the estimated maximum slope. In each full cycle over features $X_{1}, \ldots, X_{p}$, we estimate $Power_{i}$ of $X_{i}$ and use it to sort features by descending order.

{\bf Fast Kernel Smoothing Approximation}. Kernel smoothing is a popular alternative but suffers from low efficiency due to $O\left(N^{2}\right)$ complexity in general. However, a fast kernel smoothing method is proposed~\citep{Langrene:2019}, which achieves $O(N)$ complexity and with much smaller coefficient (i.e., $\ll 35$). The key idea is called fast-sum-updating: given a polynomial kernel, this method pre-computes the cumulative sum of each item in the polynomial form on evaluation points and uses these cumulative sums to perform one-shot scanning over evaluation points to complete the task. Our smoothing task takes $N$ input samples $\left(x^{(1)}, y^{(1)}\right), \ldots, \left(x^{(N)}, y^{(N)}\right)$ where $x^{(1)}, \ldots, x^{(N)}$ are also evaluation points (here we strip the feature index $i$ for simplicity). Natural cubic spline smoothing has $O(N)$ time complexity which is still expensive since it takes about $35$N operations~\citep{Silverman:1985}. We choose Epanechnikov kernel: $K(X)=3\left(1-X^{2}\right) / 4$, which is a degree-2 polynomial kernel with good theoretical property. We adopt the fast-sum-updating algorithm to approximate original cubic spline smoothing to reduce operations to $\approx  4\mathrm{N}$ almost without loss of accuracy.

\subsection{Flexibility}\label{sec4.5}
{\bf TSI's modularity}. As depicted in \figref{Figure 4}, each stage in TSI takes partial residuals as input and estimates separate parameters (i.e., corresponds to numerical, categorical, and temporal features respectively) until partial convergence (stage-level). Therefore, each stage can be performed as a standalone module, and TSI can be viewed as a framework to learn these modules iteratively. Such modularization allows us to adopt optimization techniques within each module. As aforementioned, we propose intelligence sampling and DFI to improve training efficiency in Stage1.

{\bf Learning more time series components}. TSI's modularity allows us to view Stage3 (learning on temporal features) as a classical seasonal-trend decomposition task (as depicted on the right side of \figref{Figure 7}), thus more sophisticated approaches can be adopted from literature such as STL~\citep{Cleveland:1990} or RobustSTL~\citep{Wen:2019}. Moreover, Stage3 can be extended to learn additional components such as multiple seasonal components~\citep{Cleveland:1990} or aperiodic cyclic components~\citep{Hyndman:2011}.

{\bf Tolerance to missing data}. FXAM is tolerant w.r.t. missing data in temporal features. We partition the instances into phase-$\varphi$ sets, conduct smoothing within each phase-$\varphi$ set, and learn sub-component $\boldsymbol{f}_{S_{\varphi}}$. These sub-components $\boldsymbol{f}_{S_{\varphi}}$ are further domain-merged to obtain the seasonal component $\boldsymbol{f}_S$. It is known that smoothing is good at interpolation thus it is tolerant to missing data issues.

{\bf Extension to multiple temporal features}. The previous illustration presupposes single temporal feature $T$ (for simplicity). When there are multiple temporal features $T_{1}, \ldots, T_{u}$ where $u>1$, TSI can be extended naturally by applying Stage3 for each temporal feature provided that the period of its seasonal component is given.

\section{Evaluation}
We evaluate FXAM on both synthetic and real data sets. We generate synthetic data sets to comprehensively evaluate FXAM's performance against varied data scales and data characteristics, and we use 13 representative real data sets and a case study to demonstrate the effectiveness of FXAM. 

{\bf Comparison algorithms}. We choose 3 representative algorithms for comparison: pyGAM~\citep{SerBru:2018}, EBM~\citep{Nori:2019}, and XGBoost~\citep{Chen:2015}. pyGAM is a standard implementation of GAM in python and EBM is the implementation of GA2M. We choose opaque model XGBoost as a reference for accuracy. The detailed API calls are shown in \appref{secE}.

{\bf Hardware}. All experiments are conducted on a Windows Server 2012 machine with 2.8GHz Intel Xeon CPU E5-2680 v2 and 256GB RAM. FXAM is implemented by C\#. We use the latest version of pyGAM, EBM, and XGBoost in python.

{\bf Design and metric}. We design experiments to evaluate:

    \begin{itemize}

      \item {\it Modeling}: FXAM's effectiveness in addressing one-to-many and many-to-one phenomena by varying scales of categorical or temporal features.
      
      \item {\it Training}: The performance of TSI procedure by comparing with pyGAM.
      
      \item {\it Efficiency}: FXAM vs. all competitors on training speed. 

    \end{itemize}

To measure training time of ML model used in predictive analytics, we fix the hyperparameters of each competitor algorithm beforehand. These hyperparameters are carefully tuned to achieve the best performance (details are shown in \appref{secE}). For each data set, we conduct 5-fold cross-validation and use average root-mean-square error (RMSE) to measure accuracy and we record average training time.

\subsection{Evaluation on Synthetic Data}
{\bf Synthetic data generation}. To thoroughly evaluate FXAM's performance against varied data scales / characteristics, we synthetically generate data sets by specifying a configuration which is composed of seven factors as shown in \tabref{tabl2}. Factors $1 - 3$ specify data scale, factors $4 - 6$  specify data characteristics and factor 7 specifies the difficulty level of ground-truth generation functions. The generation functions in easy mode follows standard additive models, i.e., response is the sum of contribution of each feature and then with a small random noise. The hard mode considers significant portion of feature interactions and with higher noise level (details are in \appref{secA}).

\begin{table}[width=.5\linewidth]
\caption{Seven factors for generating synthetic data sets.}\label{tabl2}
\begin{tabular*}{\tblwidth}{@{} LLLL@{} }
\toprule
ID & Factor & Value Range \\
\midrule
1 & \#records & $[10000, 500000]$ \\

2 & \#features & $[20, 200]$ \\

3 & \#total cardinalities & $[0, 2000]$ \\

4 & numerical feature ratio & $[0,1]$ \\

5 & has temporal feature & \{yes, no\} \\

6 & seasonality ratio & $[0.0,0.1]$ \\

7 & difficulty & \{easy, hard\} \\
\bottomrule
\end{tabular*}
\end{table}

\begin{figure*}[htbp]
\centering{\includegraphics[width=0.8\textwidth]{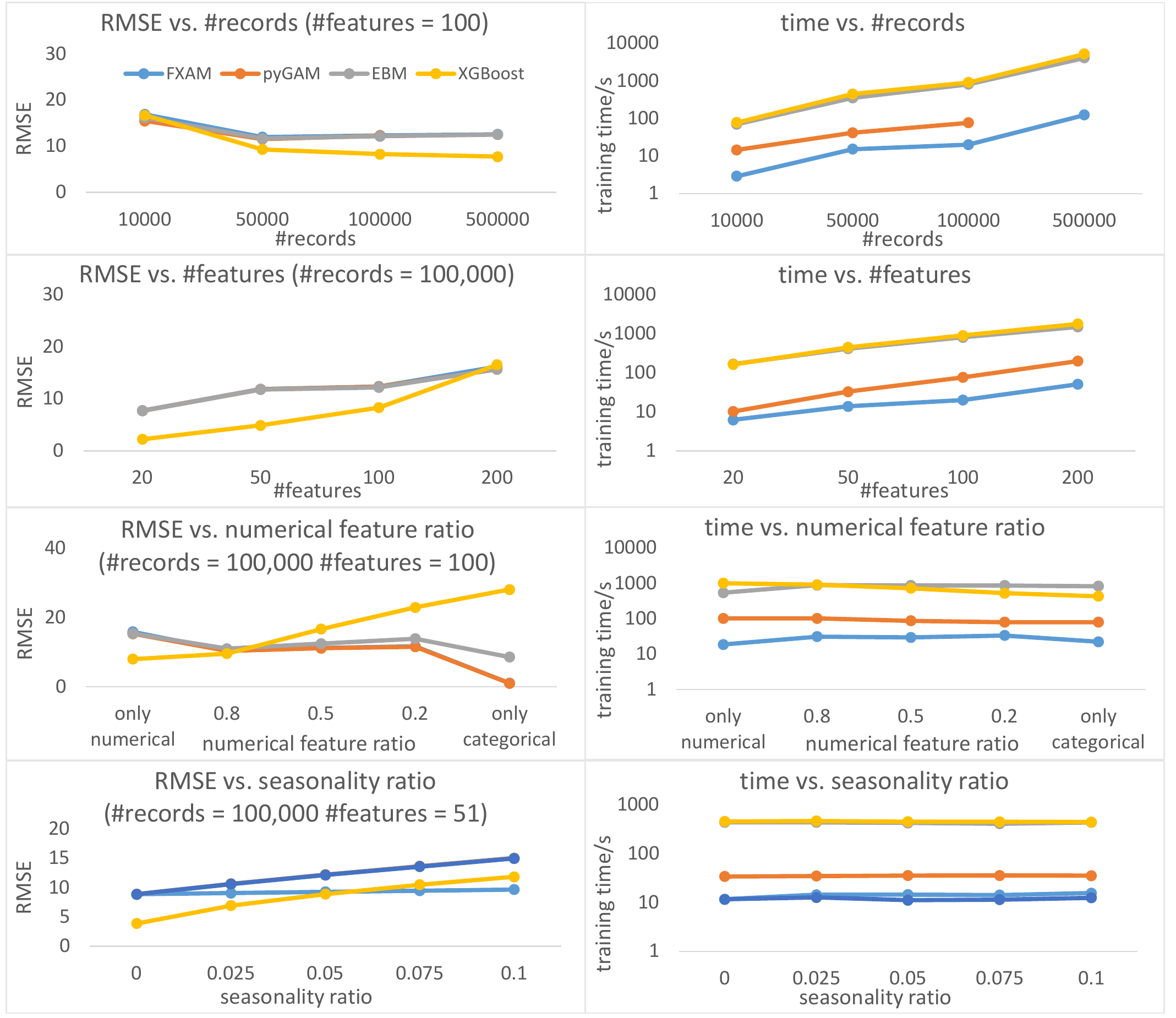}}
\caption{Evaluation on synthetic data sets. Four rows reflect different data scale / characteristic by varying: Row1: \#records $|$ Row2: \#features $|$ Row3: numerical feature ratio $|$ Row4: seasonality ratio of the temporal feature.}\label{Figure 8}
\end{figure*}

{\bf Results}. We have conducted evaluations by varying \#records, \#features, and so on. In each setup, we fix the other factors and only vary a specific one (details are in \appref{secB}). \figref{Figure 8} depicts results on hard data sets. The results of training time are using logarithmic scale. FXAM performs much better than XGBoost on accuracy and efficiency. Here only show the results on `hard' data sets. For `easy' data sets, since the ground-truth generating mechanism is with feature contributions fully untangled, thus GAM related approaches could achieve optimal accuracy, this is why XGBoost does not perform well on `easy' data sets. The complete results are shown in \appref{secD}.

{\it Addressing One-To-Many over temporal features}. The $4^{\text {th}}$ row of \figref{Figure 8} illustrates the effectiveness of FXAM on learning trends and seasonal components over temporal features. The seasonality ratio (defined as Fraction-of-Variance-Explained: \cite{Achen:1990}) is varied from $0 \%$ to $10 \%$. We also compare with a simplified version of FXAM called "FXAM\_no\_TAS", i.e., treating the temporal feature as a normal numerical feature. As seasonality ratio increases from $0 \%$ to $10 \%$, the RMSEs of all the algorithms increase except FXAM's RMSE which remains stable and achieves the highest accuracy.

{\it Addressing Many-To-One over categorical features}. In the first chart of $3^{\text {rd }}$ row of \figref{Figure 8}, the right-most data points show RMSEs when all features are categorical with total cardinality $=2,000$. Both FXAM and pyGAM achieve the smallest RMSE since they have the same regularization on categorical features. FXAM's training speed is 3 times faster than pyGAM and 10 times faster than EBM / XGBoost due to its joint learning strategy. 

{\it Efficiency and Convergence of TSI}. Results in \figref{Figure 8} generally show the performance of FXAM's training procedure TSI. In the $1^{\text {st }}$ column, XGBoost achieves the best accuracy. Meanwhile, FXAM achieves close or even better accuracy vs. pyGAM or EBM. ALL results in the $2^{\text {nd }}$ column show that FXAM achieves magnitude-order speed-up.

{\it Feasibility for interactive ML (iML)}. The typical scale of multi-dimensional data is with \#records $=100,000$, and \#features = 100. FXAM uses less than 20 seconds for training on such scale data set whereas the other algorithms cost more than 100 seconds (pyGAM) or even 1,000 seconds (EBM or XGBoost). pyGAM throws Out-Of-Memory exception when training on data sets with 500,000 records. To facilitate smooth iML experience, the machine is asked to respond within 10 seconds. Therefore, FXAM is feasible to facilitate iML in an interactive and iterative manner.

\begin{figure*}[htbp]
\centering{\includegraphics[width=0.8\textwidth]{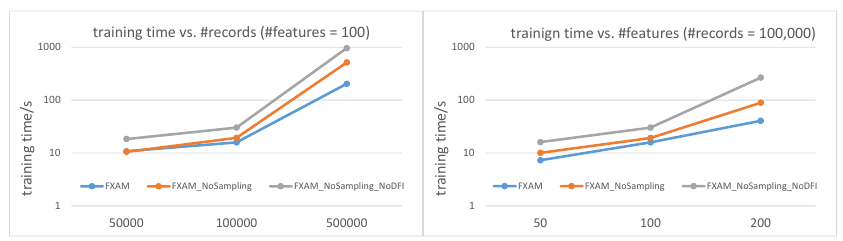}}
\caption{Ablation study of FXAM.}\label{Figure 9}
\end{figure*}

{\bf Ablation Study}. We also conduct ablation study to evaluate the efficiency improvement by using two novel techniques proposed in FXAM ( hyperparameter are shown in \appref{secC}): intelligent sampling and DFI (Dynamic Feature Iteration). As shown in \figref{Figure 9}, we compare the efficiency of FXAM among disabling sampling (FXAM\_NoSampling, orange curve), both sampling and dynamic feature iteration disabled (FXAM\_NoSampling\_NoFDI, grey curve) and original model (FXAM, blue curve). All data sets are with difficulty level = `hard'. The results show that sampling and DFI improve efficiency significantly, which confirms that sampling indeed identifies better initialization of smoothing functions, and dynamic feature iteration increases convergent speed. All three algorithms have the same RMSE because these two techniques only accelerate the convergence speed.

\subsection{Evaluation on Real Data}

We have collected 13 representative real data sets from diverse domains as shown in \tabref{tabl3}.

\begin{table}[width=.6\linewidth]
\caption{Details of real data sets.}\label{tabl3}
\begin{threeparttable}
\begin{tabular*}{\tblwidth}{@{} LLLLL@{} }
\toprule
Name        & \#Records & \#Features & Has Temporal & Domain      \\ 
\midrule
BikeShare\tnote{1}   & 17,414    & 8          & Y            & Social      \\ 
SY\tnote{1}          & 20,451    & 10         & Y            & Environment \\ 
Clif\tnote{1}        & 7,484     & 25         & Y            & Sales       \\ 
CD\tnote{1}          & 27,366    & 10         & Y            & Environment \\ 
Kickstarter\tnote{1} & 55,427    & 6          & Y            & Financial   \\ 
SH\tnote{1}          & 34,040    & 8          & Y            & Environment \\ 
CSAT\tnote{2}        & 31,779    & 11         & Y            & Computer    \\ 
Energy\tnote{3}      & 19,735    & 28         & Y            & Energy      \\ 
GZ\tnote{1}         & 32,353    & 10         & Y            & Environment \\ 
IT\tnote{2}          & 341,721   & 6          & N            & Computer    \\ 
Asteroid \tnote{1}   & 137,680   & 26         & N            & Astronomy   \\ 
Autos\tnote{1}       & 304,133   & 11         & N            & Sales       \\ 
Birth\tnote{1}       & 1,000,000 & 46         & N            & Healthcare  \\ 
\bottomrule
\end{tabular*}
 \begin{tablenotes}
        \footnotesize
        \item[1] This data set is from Kaggle https://www.kaggle.com/.  %此处加入注释*信息
        \item[2] This data set is from Microsoft Research https://www.microsoft.com/en-us/research/. %此处加入注释**信息
        \item[3] This data set is from Archive https://archive.org/. %此处加入注释**信息
      \end{tablenotes}
\end{threeparttable}
\end{table}

\begin{figure*}[]
\centering{\includegraphics[width=0.8\textwidth]{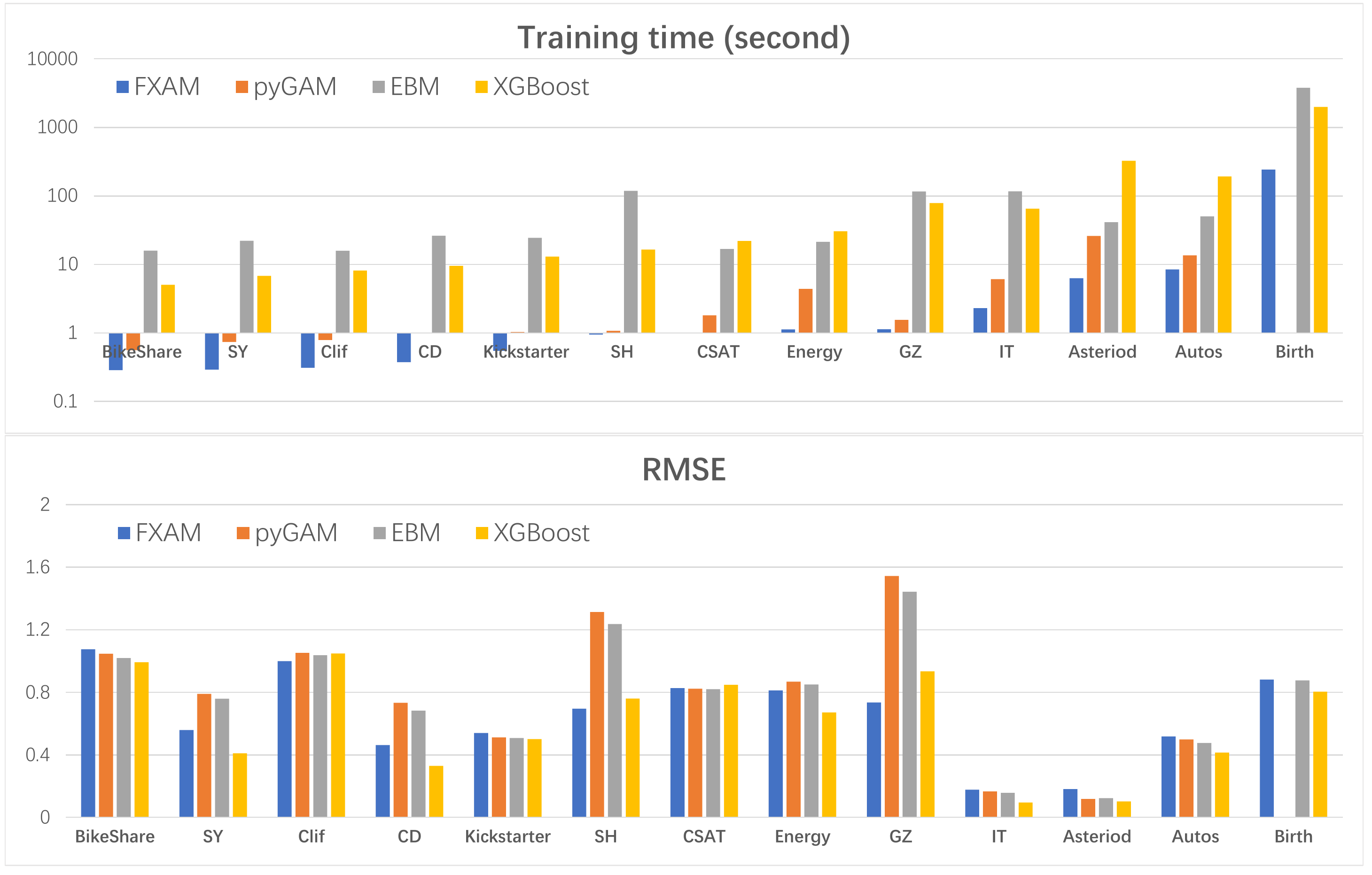}}
\caption{Training time (logarithmic scale in y-axis of top figure) and RMSE (bottom figure) on 13 real data sets.}\label{Figure 10}
\end{figure*}

{\bf Results}. As depicted in \figref{Figure 10}, compared with pyGAM and EBM, FXAM achieves the best accuracy on $6$ of $9$ data sets which have temporal features. For the left 3 data sets `BikeShare', `Kickstarter' and `CSAT', FXAM's accuracy is still very competitive. Such result reflects the effectiveness of FXAM to decompose trend and seasonal components (FXAM exhibits even better accuracy than XGBoost on three data sets `Clif', `SH', and `GZ'). Regarding training speed, FXAM is significantly faster than the other algorithms. FXAM uses $<10$ seconds to finish training on 12 out of 13 data sets except for the ultra-large data set `Birth', where FXAM uses $\approx200$ seconds but XGBoost and EBM use nearly one hour and pyGAM runs into Out-Of-Memory exception. Therefore, FXAM is suitable to facilitate iML.

{\bf Real-world case: predicting customer satisfaction compared with XGBoost+SHAP}. `CSAT' is about customer satisfaction of using an online service system. Each record corresponds to a specific customer feedback, with a satisfaction score ranging from 1 to 5. Engineers initially collect 20 features to predict the score. We build an Excel add-in to facilitate user interaction with FXAM. Engineers want to know: {\it 1) whether these features are effective to support expert decision-making; 2) how to support expert decision-making? For example, whether it is necessary to add customer service on weekends.} We show how FXAM facilitates answering these questions.

In contrast, we use SHAP~\citep{lundberg:2020} to explain the prediction results of XGBoost and the average of SHAP values of the feature is calculated as the total contribution of the feature. It should be noted that in general, each feature has a positive or negative SHAP value on each instance, and in this experiment,  the SHAP values of the feature remain positive or negative on the vast majority of instances, so there is no offset between the positive and negative contributions when calculating the average.

\begin{figure*}[]
\centering{\includegraphics[width=0.9\textwidth]{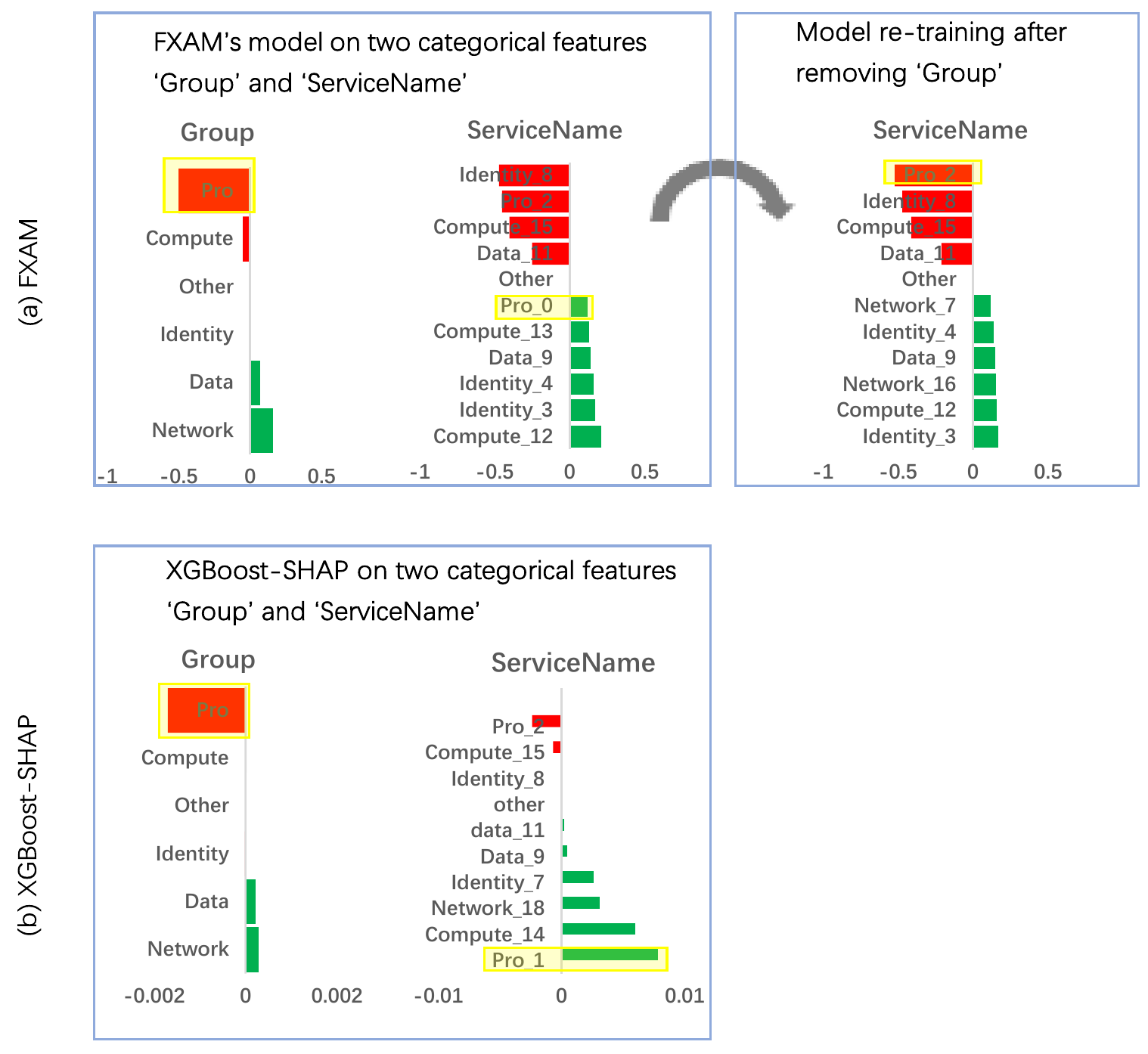}}
\caption{Model interpretability comparison on categorical features (Combined with domain knowledge, the results show that FXAM accurately detects the positive contribution of $Pro$, while XGBoost+SHAP mistakenly identifies the contribution of $Pro$ as negative).}\label{Figure 11}
\end{figure*}

{\it FXAM is trustworthy enough to support expert decisions}. The contributions of features output by a trustworthy model should be consistent with the ground truth. The following takes $Pro$ as an example to illustrate the trustworthiness of FXAM. \figref{Figure 11}(a) depicts the FXAM model on categorical features $Group$ and $ServiceName$. $ServiceName$ is the subdivision of $Group$ to indicate specific services within each $Group$ (each service name takes its corresponding group name as prefix). FXAM reports a positive contribution from $Pro \_ 0$ (box selected) on $SeriviceName$. However, due to domain knowledge, services within $Pro$ are more likely to have negative contributions. A feasible explanation is that $Pro$ from feature $Group$ absorbs the most negative contribution (1 $^{\text {st}}$ column in \figref{Figure 11}(a)). Considering the dependency between $Group$ and $ServiceName$, engineers delete $Group$ and then re-train FXAM. In seconds, FXAM returns an updated model which is verified to be meaningful ( $3^{\text {rd }}$ column of \figref{Figure 11}(a)). Following such interactive analysis flow, engineers eventually select 11 features to predict score and the corresponding model is verified to be trustworthy.

As shown in \figref{Figure 11}(b), when explaining XGBoost with SHAP, $Pro$ shows the same contradictory results on $Group$ and $Servicename$. But it also shows that $Pro$ tends to have positive impact, which contradicts our domain knowledge.

\begin{figure}[]
\centering{\includegraphics[width=0.8\textwidth]{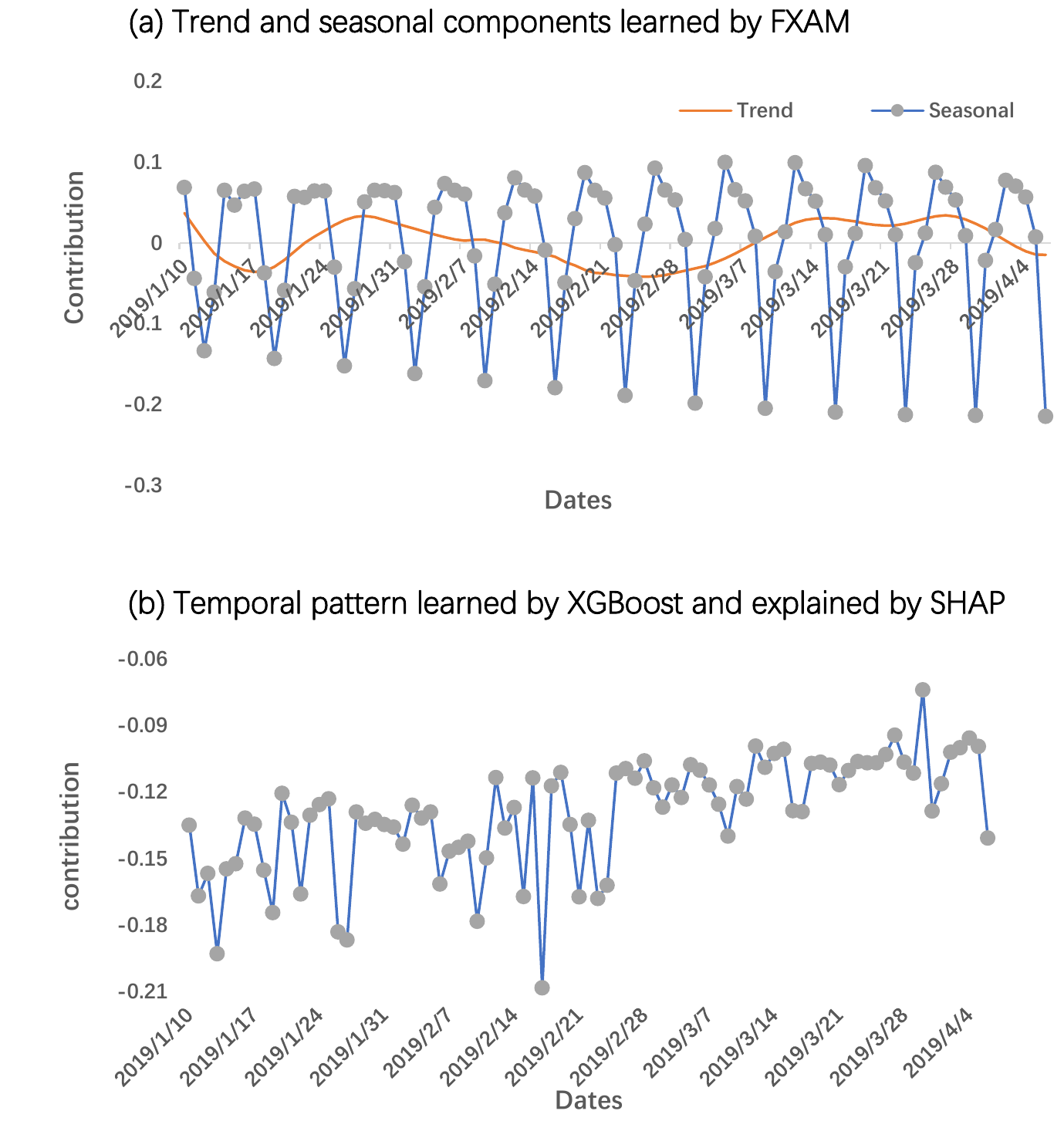}}
\caption{Model interpretability comparison on feature $Date$ (FXAM can disentangle the seasonality and trend contributions of temporal features, while they are entangled in XGBoost+SHAP 's results).}\label{Figure 12}
\end{figure}

{\em Customer service should be added on weekends, based on FXAM's interpretability on $Date$}. In \figref{Figure 12}(a), FXAM identifies a significant seasonal component $\boldsymbol{f}_S$ (blue curve) and a trend component $\boldsymbol{f}_T$ (orange curve) on feature $Date$. In $\boldsymbol{f}_S$, weekly repeating pattern is exhibited and each Saturday (i.e., valley points) has the most negative contribution within a period. There is no clear upward or downward trending in $\boldsymbol{f}_T$. Thus, a stable user behavior is exhibited such that customer satisfaction becomes worse during the weekend. Moreover, the amplitude of $\boldsymbol{f}_S$ is growing, which suggests the urgency to allocate more customer service on the weekend.

The results of XGBoost+SHAP is shown in \figref{Figure 12}(b). The trend and seasonality are mixed together, which cannot be effectively analyzed and cannot help answer the second question. 

\section{Conclusion}
We have proposed FXAM, which extends GAM's modeling capability with a unified additive model for numerical, categorical, and temporal features. FXAM addresses the challenges introduced by one-to-many and many-to-one phenomena, which are commonly appeared in predictive analytics. FXAM conducts a novel training procedure called TSI (Three-Stage Iteration). We prove that TSI is guaranteed to converge and the solution is globally optimal. We further propose two novel techniques to speed up FXAM's training algorithm to meet the needs of interactive ML. Evaluations have verified that FXAM remarkably outperforms the existing GAMs regarding training speed and modeling categorical or temporal features.

The success of FXAM's real-world adoption has demonstrated the importance of interactive and interpretable ML, which are also the main design goals of FXAM. One future direction is to extend FXAM to support not only main effects (i.e., the univariate shape functions in this paper) but also pairwise or higher-order interactions such as GA2M~\citep{Lou:2013}. It is known that learning interactions will introduce significantly higher computational costs, and the high-efficiency training speed of FXAM is clearly a benefit. 

\section*{CRediT authorship contribution statement}
{\bf Yuanyuan Jiang}: Conceptualization, Methodology, Software, Validation, Formal analysis, Investigation, Data curation, Writing – original draft, Writing – review \& editing. {\bf Rui Ding}: Conceptualization, Methodology, Software, Validation, Formal analysis, Investigation, Data curation, Writing – original draft, Writing – review \& editing. {\bf Tianchi Qiao}: Conceptualization, Methodology, Software, Validation, Investigation. {\bf Yunan Zhu}: Conceptualization, Methodology, Software, Validation, Investigation. {\bf Shi Han}: Conceptualization, Writing – review \& editing. {\bf Dongmei Zhang}: Conceptualization, Writing – review \& editing.

\section*{Declaration of Competing Interest}
The authors declare that they have no known competing financial interests or personal relationships that could have appeared to influence the work reported in this paper.

\section*{Acknowledgements}
This research did not receive any specific grant from funding agencies in the public, commercial, or not-for-profit sectors.

\section*{Data availability}
Data will be made available on request.

% Uncomment and use as the case may be
%\begin{theorem} 
%\end{theorem}

% Uncomment and use as the case may be
%\begin{lemma} 
%\end{lemma}


\printcredits

%\bibliographystyle{cas-model2-names}
\bibliographystyle{apalike}
\bibliography{manuscript}



\clearpage
\begin{appendices}

\section{Generation Details of Synthetic Data}\label{secA}

\subsection{Generation for Numerical Features}
{\bf Easy Mode}. The numerical features are generated based on three univariate functions as follows:

\begin{enumerate}
  \item $f(X)=A_{1} X$
  
  \item $g(X)=A_{2} X^{2}+A_{3} X$
  
  \item $h(X)=A_{4} \sin \left(A_{5} X+A_{6}\right)$ 
\end{enumerate}

For a specific numerical feature, we first randomly choose one of the three functions by probabilities: $0.3: 0.3: 0.4$ (w.r.t. $f$, $g$ and $h$ respectively). $A_{1}, A_{3}, A_{4}$ are random variables that are uniformly and independently drawn from $[-2,2]$. $a_{2}$ is drawn uniformly from $[-1,1]$. $A_{5}$ is drawn uniformly from $[0,6 \pi]$, and $A_{6}$ is drawn uniformly from $[-0.5,0.5]$.

Once the coefficients $A_{1}, \ldots, A_{6}$ are set, our generator generates each record with variable $X$ drawn uniformly from $[0,10]$.

The final response is the total sum of each function and additionally with a random noise $\epsilon: \epsilon$ follows normal distribution with $E(\epsilon)=0$, and its variance is adjusted based on generated data so that the $\operatorname{Var}(\epsilon) / {T S S}=0.1 \%$.

{\bf Hard Mode}. Besides the three univariate functions, we include two additional two-variable functions $I_{1}$ and $I_{2}$ to indicate feature interactions:

    \begin{itemize}

      \item $I_{1}(X_1, X_2)=B_{1} X_1 X_2+B_{2} X_1+B_{3} X_2$

      \item $I_{2}(X_1, X_2)=B_{4} \cos \left(B_{5} X_1 X_2+B_{6} X_1+B_{7} X_2+B_{8}\right)$

    \end{itemize}
    
In hard mode, for a specific numerical feature, we first randomly choose one of the five functions $\left\{f, g, h, I_{1}, I_{2}\right\}$ by probabilities: $0.1: 0.1: 0.2: 0.2: 0.4$ accordingly. If the function is drawn to be either $I_{1}$ or $I_{2}$, we will use two numerical features to generate their contributions to the response.

Coefficients $B_{1}, B_{2}, B_{3}, B_{4}$ are random variables that are uniformly and independently drawn from $[-2,2]$. $B_{5}, B_{6}, B_{7}$ are drawn uniformly from $[0,4 \pi]$, and $B_{8}$ is drawn uniformly from $[-0.5,0.5]$

Once the coefficients are set, if the function is either $I_{1}$ or $I_{2}$, the two variables are drawn uniformly and independently $X_1 \sim[0,10], X_2 \sim[0,10]$ to generate each record.

The final response is the total sum of each function and additionally with a random noise $\epsilon$. We adjust the interaction items to assure they contribute $60 \% - 70 \%$ to final response (w.r.t. Fraction of Variance Explained by interaction items). $\epsilon$ follows normal distribution with $E(\epsilon)=0$, and its variance is adjusted based on generated data so that the $\operatorname{Var}(\epsilon) / {T S S}=0.5 \%$. Thus, the noise level for hard mode is five times larger than it for easy mode

\subsection{Generation for Categorical Features}
For each categorical feature, its cardinality is set uniformly from integers in [2, $MaxCardinality$]. $MaxCardinality$ is a configuration parameter with value ranging from 10 to 38 (so the average cardinality is from 6 to 20).

The contribution of each specific categorical value $Z_{i}$ is $\beta\left(Z_{i}\right)$, called weight, and $\beta\left(Z_{i}\right) \sim[0,15]$ which is drawn independently and uniformly.

\subsection{Generation for Temporal Feature}
We inject seasonality components into data by considering a temporal feature with the form:
$$
f_{TS}(T)=V_{1} \sin \left(\frac{2 \pi T}{10}+V_{2}\right)
$$
Here $V_{2} \sim[-5,5]$ and $V_{1}$ is used to control the ratio of seasonality components (w.r.t. its influence on final response).

$T$ is the discrete time, so it is an integer randomly drawn from $[1,200]$. \\
\\
\\
\\
\section{Configurations for Evaluation on Synthetic Data}\label{secB}

\subsection{Varying \# Records}
Here we set $MaxCardinality =10$ per each categorical feature, so the expectation of total cardinality is 120 as shown in \tabref{tabl4}.

\newcounter{Stable}
\setcounter{Stable}{1}
\renewcommand{\thetable}{B.\arabic{Stable}}

\begin{table}[width=.5\linewidth]
\caption{Varying \# Records.} \label{tabl4}
\begin{tabular*}{\tblwidth}{@{} LLLL@{} }
\toprule
ID & Factor & Value Range \\
\midrule
1 & \#records & $[10000,500000]$ \\

2 & \#features & 100 \\

3 & \#total cardinalities & 120 \\

4 & numerical feature ratio & $0.8$ \\

5 & has temporal feature & no \\

6 & difficulty & \{easy, hard\} \\
\bottomrule
\end{tabular*}
\addtocounter{Stable}{1}
\end{table}

\subsection*{B.2. Varying \# Features}
Here we set $MaxCardinality =10$ per each categorical feature as shown in \tabref{tabl5}.

\setcounter{Stable}{2}
\renewcommand{\thetable}{B.\arabic{Stable}}

\begin{table}[width=.5\linewidth]
\caption{Varying \# Features.}\label{tabl5}
\begin{tabular*}{\tblwidth}{@{} LLLL@{} }
\toprule
ID & Factor & Value Range \\
\midrule
1 & \#records & 100,000 \\

2 & \#features & $[20,200]$ \\

3 & \#total cardinalities & $[24,240]$ \\

4 & numerical feature ratio & $0.8$ \\

5 & has temporal feature & no \\

6 & difficulty & easy, hard \\
\bottomrule
\end{tabular*}
\addtocounter{Stable}{2}
\end{table}

\subsection{Varying Numerical Feature Ratio}
Here we set $MaxCardinality=38$ per each categorical feature as shown in \tabref{tabl6} so that the expectation of total cardinality is

$0$: when numerical feature ratio $=1$;

$2000:$ when numerical feature ratio $=0$ \\
\\

\setcounter{Stable}{3}
\renewcommand{\thetable}{B.\arabic{Stable}}

\begin{table}[width=.5\linewidth]
\caption{Varying Numerical Feature Ratio.}\label{tabl6}
\begin{tabular*}{\tblwidth}{@{} LLLL@{} }
\toprule
ID & Factor & Value Range \\
\midrule
1 & \#records & 100,000 \\

2 & \#features & 100 \\

3 & \#total cardinalities & $[0,2000]$ \\

4 & numerical feature ratio & $[0,1]$ \\

5 & has temporal feature & no \\

6 & difficulty & \{easy, hard\} \\
\bottomrule
\end{tabular*}
\addtocounter{Stable}{3}
\end{table} 

\subsection{Varying Seasonality Ratio from Temporal Feature}
Here we set $MaxCardinality=10$ per each categorical feature as shown in \tabref{tabl7}.

\setcounter{Stable}{4}
\renewcommand{\thetable}{B.\arabic{Stable}}

\begin{table}[width=.5\linewidth]
\caption{Varying Seasonality Ratio from Temporal Feature.}\label{tabl7}
\begin{tabular*}{\tblwidth}{@{} LLLL@{} }
\toprule
ID & Factor & Value Range \\
\midrule
1 & \#records & 100,000 \\

2 & \#features & 51 \\

3 & \#total cardinalities & 60 \\

4 & numerical feature ratio & $40 / 51$ \\

5 & has temporal feature & yes \\

6 & difficulty & \{easy, hard\} \\
\bottomrule
\end{tabular*}
\addtocounter{Stable}{4}
\end{table}

\section{Ablation Study}\label{secC}
Ablation study mainly evaluates two novel techniques of FXAM: intelligent sampling and dynamic feature iteration, and the hyperparameters are shown in \tabref{tabl8} and\tabref{tabl9} respectively. Since these two techniques are applied for numerical features, thus in this study, we only generate data sets with pure numerical features.

\setcounter{Stable}{1}
\renewcommand{\thetable}{C.\arabic{Stable}}

\begin{table}[width=.5\linewidth]
\caption{hyperparameters in Study 1.}\label{tabl8}
\begin{tabular*}{\tblwidth}{@{} LLLL@{} }
\toprule
ID & Factor & Value Range \\
\midrule
1 & \#records & $\{50,000,100,000,500,000\}$ \\

2 & \#features & 100 \\

3 & \#total cardinalities & 0 \\

4 & numerical feature ratio & 1 \\

5 & has temporal feature & no \\

6 & difficulty & hard \\
\bottomrule
\end{tabular*}
\addtocounter{Stable}{1}
\end{table}

\setcounter{Stable}{2}
\renewcommand{\thetable}{C.\arabic{Stable}}

\begin{table}[width=.5\linewidth]
\caption{hyperparameters in Study 2.}\label{tabl9}
\begin{tabular*}{\tblwidth}{@{} LLLL@{} }
\toprule
ID & Factor & Value Range \\
\midrule
1 & \#records & 100,000 \\

2 & \#features & $\{50,100,200\}$ \\

3 & \#total cardinalities & 0 \\

4 & numerical feature ratio & 1 \\

5 & has temporal feature & no \\

6 & difficulty & hard \\
\bottomrule
\end{tabular*}
\addtocounter{Stable}{2}
\end{table}

\section{Complete Results on Synthetic Data}\label{secD}
In paper, we only present results on `hard' data sets. Here the first two columns are additional results on `easy' data sets as shown in \figref{Figure 13}. FXAM and other related approaches perform much better than XGBoost on accuracy and efficiency for `easy' data set.

\newcounter{Sfigure}
\setcounter{Sfigure}{1}
\renewcommand{\thefigure}{D.\arabic{Sfigure}}

\begin{figure*}[ht]
\centering{\includegraphics[width=\textwidth]{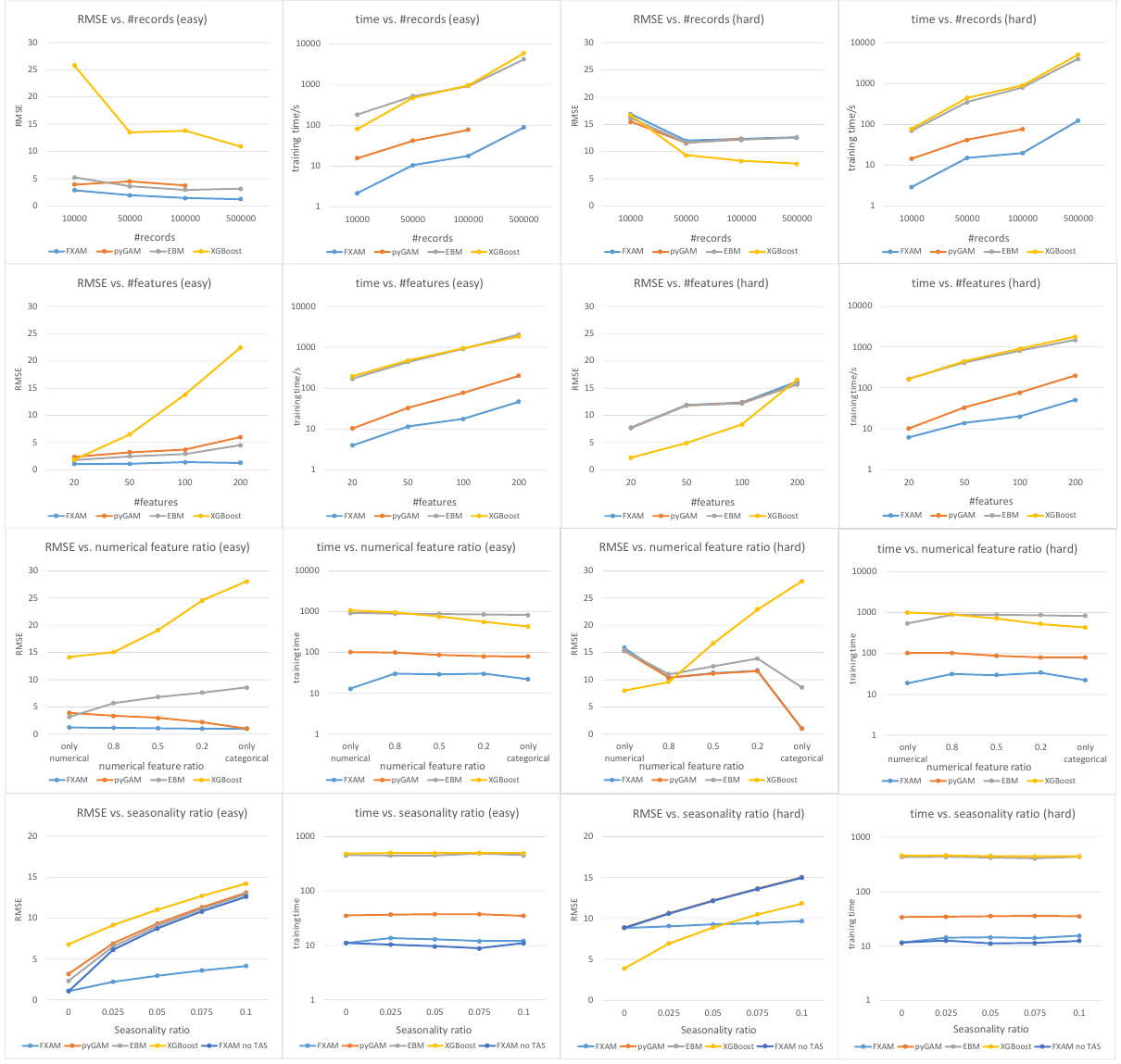}}
\caption{Performance comparison from different perspectives by varying. Row1: \#records $|$ Row2: \#features $|$ Row3: numerical feature ratio $|$ Row4: seasonality ratio of temporal feature.}\label{Figure 13}
\addtocounter{Sfigure}{1}
\end{figure*}

\section{API Calls for XGBoost, EBM, mgcv and pyGAM}\label{secE}
To make fair comparison with FXAM, for categorical features, we conduct the same one-hot encoding and apply it to all the competing algorithms. 

Below are the detailed API calls that we choose for comparison, the hyperparameters are carefully tuned to make the result accurate and as fast as possible. 

\subsection{EBM}

We use the python code from \citet{Nori:2019} for evaluation. The detailed API calling is: ExplainableBoostingRegressor ($\mathrm{n}$\_estimators $=16$, learning\_rate $=0.01$, max\_tree\_splits $=2$, (default parameters) n\_jobs $=1$)

\subsection{pyGAM}
We use the python code from \citet{SerBru:2018} for evaluation. The detailed API calling is

\textcolor[RGB]{4,227,79}{
/* For pyGAM, we choose to fit categorical feature with smoothing function type:}

\textcolor[RGB]{4,227,79}{"f()", i.e. factor term; we choose to fit numerical feature with smoothing function type:}

\textcolor[RGB]{4,227,79}{"s()", i.e. spline term.}

\textcolor[RGB]{4,227,79}{Therefore, terms is a list pre-generated based on feature type, which is used to indicate which type of smoothing function is selected for the corresponding feature. */
}

LinearGAM(terms, max\_iter $=100$, tol $=1 \mathrm{e}-4$ )

\subsection{XGBoost}
We choose three typical versions of parameters to run XGBoost, which are 1) fast, 2) mild, and 3) slow. The "fast" version is with fast training speed but accuracy is low, and the "slow" version is with good accuracy but training speed is low. "mild" is a set of parameters which we carefully tuned; thus it is a good balance, which is the version used in our evaluation. The results of three typical versions are shown in \figref{Figure 14}.

\setcounter{Sfigure}{1}
\renewcommand{\thefigure}{E.\arabic{Sfigure}}

\begin{figure}[ht]
\centering{\includegraphics[width=\textwidth]{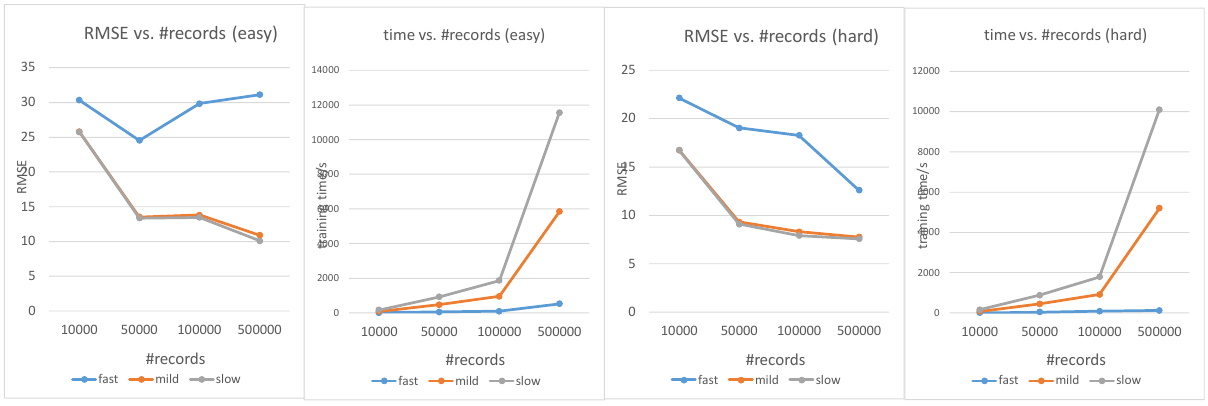}}
\caption{Performance results of XGBoost for three different versions.}\label{Figure 14}
\addtocounter{Sfigure}{1}
\end{figure}

For instance, below we show our experiments for three version of parameters to call XGBoost. You can see that "mild" achieves very close accuracy with "slow" but its time is much faster.

{\bf Fast.}
n\_estimators $=100$, learning rate $=0.3$, max\_depth $=6$, min\_child\_weight=1

{\bf Mild.}
n\_estimators $=500$, learning rate $=0.3$, max\_depth $=7$, min\_child weight=5

{\bf Slow.}
n\_estimators $=1000$, learning rate $=0.1$, max\_depth $=7$, min\_child\_weight=5

\end{appendices}
\end{document}